\documentclass{article}

\usepackage{microtype}
\usepackage{graphicx}
\usepackage{subcaption}
\usepackage{booktabs} %

\usepackage{hyperref}

\usepackage[preprint]{icml2026}

\usepackage{amsmath}
\usepackage{amssymb}
\usepackage{mathtools}
\usepackage{amsthm}

\usepackage{pifont}
\usepackage{adjustbox}
\usepackage{multicol}
\usepackage{multirow}
\usepackage{makecell}
\usepackage{enumitem}

\usepackage{contour}
\contourlength{0.5pt}

\newcommand{\cmark}{\ding{51}}
\newcommand{\xmark}{\ding{55}}

\usepackage[capitalize,noabbrev]{cleveref}

\theoremstyle{plain}

\theoremstyle{definition}

\theoremstyle{remark}

\usepackage[textsize=tiny]{todonotes}

\icmltitlerunning{No time to train! Training-Free Reference-Based Instance Segmentation}

\begin{document}

\twocolumn[
  \icmltitle{No time to train! Training-Free Reference-Based Instance Segmentation}

  \icmlsetsymbol{equal}{*}
  \icmlsetsymbol{phd}{\smash{$\dagger$}}

  \begin{icmlauthorlist}
    \icmlauthor{Miguel Espinosa}{equal,edi}
    \icmlauthor{Chenhongyi Yang}{equal,meta,phd}
    \icmlauthor{Linus Ericsson}{gla}
    \icmlauthor{Steven McDonagh}{edi}
    \icmlauthor{Elliot J. Crowley}{edi}
  \end{icmlauthorlist}

  \icmlaffiliation{edi}{School of Engineering, University of Edinburgh, Scotland, UK}
  \icmlaffiliation{gla}{School of Computing Science, University of Glasgow, Scotland, UK}
  \icmlaffiliation{meta}{Meta}

  \icmlcorrespondingauthor{Miguel Espinosa}{miguel.espinosa@ed.ac.uk}

  \icmlkeywords{Machine Learning, ICML}

  \vskip 0.3in
]

\printAffiliationsAndNotice{\icmlEqualContribution \textsuperscript{\smash{$\dagger$}}Work done when Chenhongyi was a PhD student at the School of Engineering, University of Edinburgh.\\}

\begin{abstract}

The performance of image segmentation models has %
historically been constrained by the high cost of collecting %
large-scale annotated data. 
The Segment Anything Model (SAM) alleviates this original problem %
through a promptable, semantics-agnostic, segmentation paradigm and yet %
still requires manual visual-prompts or complex domain-dependent prompt-generation rules to process a new image.
Towards %
reducing this new burden, our work investigates the task of %
object segmentation when provided with, alternatively, only a small set of reference images. 
Our key insight is to leverage strong semantic priors, as learned by foundation models, to identify corresponding regions between a reference and a target image. 
We find that correspondences enable automatic generation of instance-level segmentation masks for downstream tasks and 
instantiate our ideas via a multi-stage, training-free method incorporating (1) memory bank construction; (2) representation aggregation and (3) semantic-aware feature matching. Our experiments show significant improvements on segmentation metrics, leading to state-of-the-art performance on COCO FSOD (36.8\% nAP), PASCAL VOC Few-Shot (71.2\% nAP50) and outperforming existing training-free approaches on the Cross-Domain FSOD benchmark (22.4\% nAP). 

\texttt{Website:}
\url{https://miquel-espinosa.github.io/no-time-to-train}

\end{abstract}

\section{Introduction}
\label{sec:intro}

{\begin{figure*}[t]
  \centering
  \includegraphics[width=\linewidth]{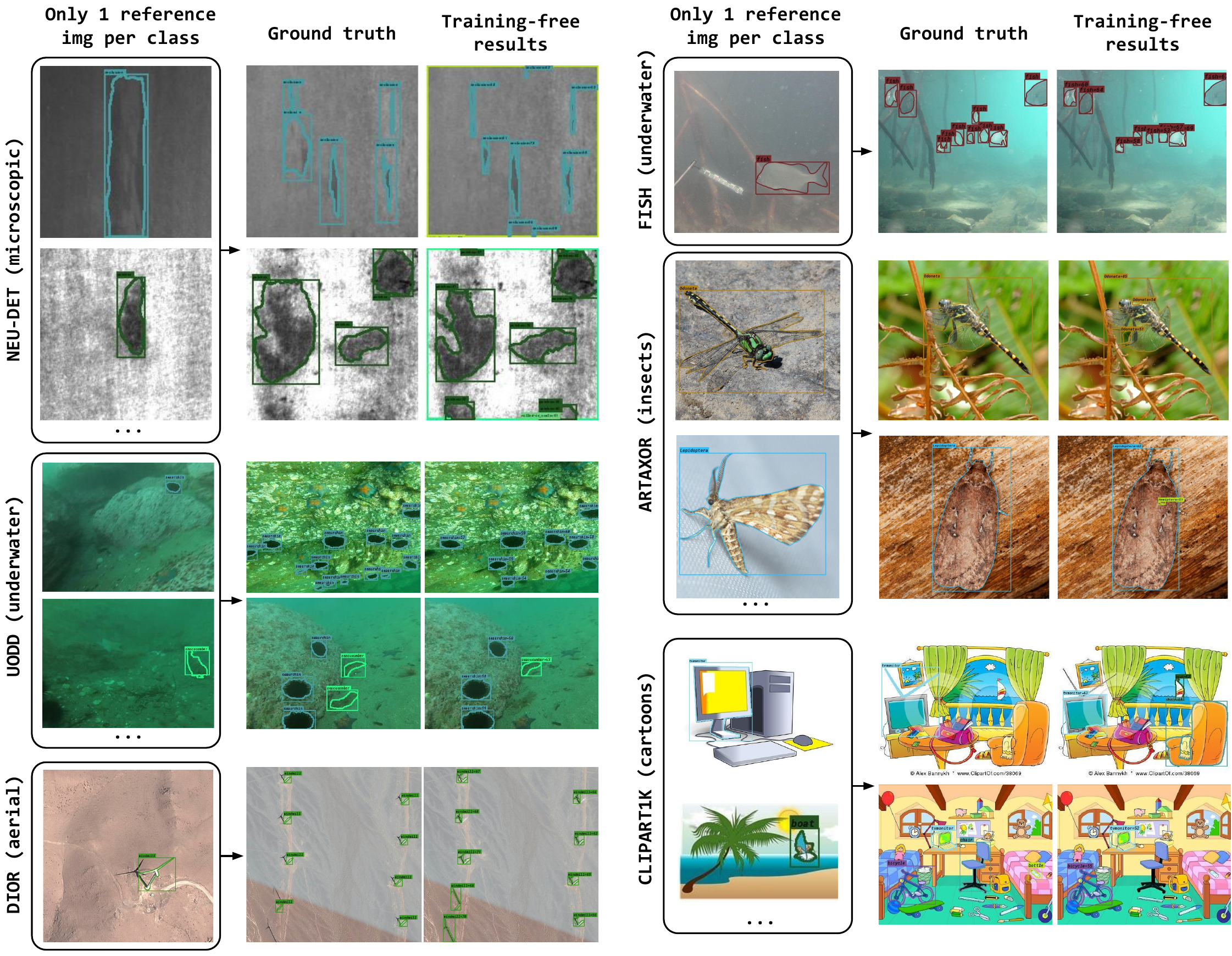}
  \caption{\textbf{Cross-domain \underline{1-shot} segmentation results using our training-free method on CD-FSOD benchmark}. Our method directly evaluates on diverse datasets without any fine-tuning, using frozen SAMv2 and DINOv2 models. The reference set contains a single example image per class. The model then segments the entire target dataset based on the reference set. Results show: (1) generalization capabilities to out-of-distribution domains (e.g., underwater images, cartoons, microscopic textures); (2) state-of-the-art performance in 1-shot segmentation without training or domain adaptation; (3) limitations in cases with ambiguous annotations or highly similar classes (e.g., ``harbor" vs. ``ships" in DIOR). Best viewed when zoomed in. Ablation studies further investigate the variance associated with the selection of reference images. See Appendix~\ref{app:cdfsod} for more visualisations.}
  \label{fig:cdsod-results}
  \vspace{-2mm}
\end{figure*}
}

It is well understood that collecting large-scale annotations for segmentation tasks is a costly and time-consuming process~\cite{benenson2019large, papadopoulos2021scaling,lvis}. Recent advances in promptable segmentation frameworks~\cite{zou2023seem, Painter, t_rex2, SegGPT, pan2023tap,liu2024grounding}, epitomised by the Segment Anything Model (SAM)~\cite{sam, sam2}, have significantly reduced manual effort by enabling high-quality mask generation using simple geometric prompts such as points, boxes or rough sketches.
While this represents a substantial advancement in reducing manual effort, these masks lack semantic awareness 
\cite{shin2024towards,no_samantics,han2023boosting,ji2024segment} and require either manual intervention or complex, domain-specific prompt-generation pipelines to function autonomously e.g.\ in medical imaging~\cite{MedSAM,Lei2023medlam,ZHANG2024108238,zhang2023towards}, agriculture~\cite{CARRARO2023100367,tripathy2024investigating}, remote sensing~\cite{osco2023segment,ma2024sam,wang2023samrs}). 
Relying on manual prompts for each image limits their scalability (especially for large datasets or scenarios requiring automatic processing), while relying on domain-constrained automated pipelines restricts the ability to generalise to cross-domain scenarios.

Reference-based instance segmentation \cite{ifsrcnn,imtfa,meta-faster-rcnn} offers a promising solution to this challenge by using %
a small set of annotated reference images to guide the segmentation of a large set of target images.
This %
idea has the potential to enable cheap, quick and automatic annotation of datasets %
where such labelling is expensive, time-consuming and requires expertise knowledge~\cite{benenson2019large, papadopoulos2021scaling}.
Unlike slow manual prompting~\cite{hu2024leveraging}, %
using reference images
can incorporate semantic understanding directly from examples, %
and is thus well-suited for automated segmentation tasks.
Despite promising results, we observe that existing reference-based segmentation methods %
often require fine-tuning on novel classes and this raises a set of well understood concerns that include task-specific data requirements, overfitting and domain shift. %
We conjecture that a prospective alternative approach to guiding reference-based instance segmentation involves reusing the general purpose capabilities of vision foundation models~\cite{dino,clip,dinov2,sam,sam2}. 

Several works~\cite{Liu2023May_matcher,sun2024vrp,SegGPT} have attempted to combine pretrained models for reference-based segmentation, e.g.~Matcher~\cite{Liu2023May_matcher} which integrates DINOv2 with SAM for semantic segmentation tasks. However, these methods face several limitations. Firstly, they rely on %
computationally expensive distance metrics (Earth Mover’s Distance), and complex thresholding mechanisms, which significantly slow down inference. Secondly, they are not suited for instance-level segmentation tasks, struggling with fine-grained discrimination in complex multi-object scenes. In fact, the instance segmentation setting presents unique challenges---how do we handle occlusions, scale variations, ambiguous object boundaries and varying image quality, all with just a few reference images---and they should be tackled thoughtfully.
Effectively combining foundation models, without significant finetuning, remains a large challenge~\cite{no_samantics}, particularly when %
attempting to leverage %
generalisation capabilities of semantic ViT backbones (e.g.~DINOv2), to achieve precise localisation~\cite{devit}.
These observations highlight that naively composing existing foundation models is insufficient for effective instance-level matching; achieving strong performance requires a dedicated system-level design, which motivates the training-free method we develop in this work. Code is made available.

We propose a training-free three-stage method: (1) constructing a memory bank of category-specific features, (2) refining feature representations via two-step aggregation, and (3) performing inference through feature matching and a novel semantic-aware soft merging strategy. This results in a training-free, high-performing framework that achieves significant
gains on established datasets. Furthermore, our approach maintains its effectiveness across diverse domains with fixed hyperparameters, making it accessible for a wide range of applications.
Our contributions are:
\begin{itemize}[noitemsep=2pt, topsep=0pt]

\item We propose a training-free method that effectively integrates semantic-agnostic segmentation mask proposals with fine-grained semantics for reference-based instance segmentation.

\item We introduce a novel three-stage framework for instance segmentation with vision foundation models, addressing key integration challenges with (1) memory bank construction, (2) two-step feature aggregation, and (3) feature matching with semantic-aware soft merging.

\item Our method achieves state-of-the-art performance on COCO-FSOD, PASCAL-FSOD, and CD-FSOD benchmarks, demonstrating strong generalisation across diverse datasets under fixed hyperparameter settings, without the need for intermediate fine-tuning.

\end{itemize}

\vspace{-2mm}
\section{Related Work}
\label{sec:rel_works}
\vspace{-1mm}

\noindent\textbf{Reference-based Instance Segmentation} aims to segment individual objects within an image, distinguishing between instances of the same category~\cite{chen2019tensormask, li2017fully}. Traditional approaches, such as Mask R-CNN~\cite{maskrcnn}, use region proposals with convolutional networks to predict instance masks, while transformer-based models like DETR~\cite{detr} and Mask2Former~\cite{mask2former} integrate global context through self-attention mechanisms. These methods have demonstrated success in standard instance segmentation tasks by leveraging large, labeled datasets~\cite{coco, lvis}. Reference-based instance segmentation extends these tasks to handle novel categories with limited labeled examples. Early works \cite{imtfa, ifsrcnn} adapted Mask R-CNN by introducing instance-level discriminative features \cite{imtfa} or uncertainty-guided bounding box prediction \cite{ifsrcnn}. More recent works have unified segmentation tasks into in-context learning frameworks \cite{Painter, SegGPT, t_rex2};{\color{blue}~\cite{dinox}}, involving an expensive pretraining phase on a wide range of segmentation tasks, including instance segmentation \cite{Painter, SegGPT};{\color{blue}~\cite{dinox}} and using contrastive pretraining to integrate visual and textual prompts \cite{t_rex2}. Despite these advancements, reference-based instance segmentation remains challenging due to the lack of labeled data, the complexity of multi-instance scenarios, limited generalisation across domains for specialist models, reference image ambiguity, and reliance on predefined class labels \cite{zang2021fasa}. Additionally, reusing frozen backbones not originally pretrained for instance segmentation remains a challenge \cite{devit, no_samantics}. Our method effectively reuses two existing frozen vision foundation models---none of which was trained for reference-based instance segmentation task--- to tackle reference-based instance segmentation without additional training, while also generalising
to unusual domains.

\noindent\textbf{Vision Foundation models} have revolutionised computer vision by learning strong, pretrained representations, transferable across diverse tasks. CLIP \cite{clip} and DINO models \cite{dino, dinov2} exemplify this trend, using contrastive learning to align visual and textual representations, and learning robust image embeddings from unlabeled data, respectively. These models have been widely adopted for downstream tasks, including open-vocabulary detection \cite{shin2024towards, han2023boosting} and semantic segmentation \cite{time, ziegler2022self}. However, CLIP struggles with detailed spatial reasoning, while DINOv2, despite capturing fine-grained semantics, produces low-resolution feature maps. The Segment Anything Models (SAM, SAM2) \cite{sam,sam2} are a notable addition to this category, trained on an extensive, category-agnostic dataset (SA-1B) \cite{sam}. While SAM excels in generating segmentation masks with minimal input (e.g., points, bounding boxes), it lacks inherent semantic understanding \cite{shin2024towards,no_samantics,han2023boosting,ji2024segment}. Efforts to bridge this gap
include pairing SAM with language models \cite{lai2023lisa}, diffusion models \cite{diffews}, or fine-tuning it on labeled datasets like COCO \cite{coco} and ADE20K \cite{semanticsam}. However, these adaptations often result in complex pipelines or limited scalability. Furthermore, SAM’s semantic-agnostic nature poses limitations in scenarios requiring class differentiation at instance-level. Our work combines the complementary strengths of DINOv2 and SAM without requiring finetuning. By aggregating and matching features from multiple references, we enable high-precision, training-free instance segmentation, achieving state-of-the-art results on diverse few-shot benchmarks.

{\begin{figure*}[t]
  \centering
  \includegraphics[width=\linewidth]{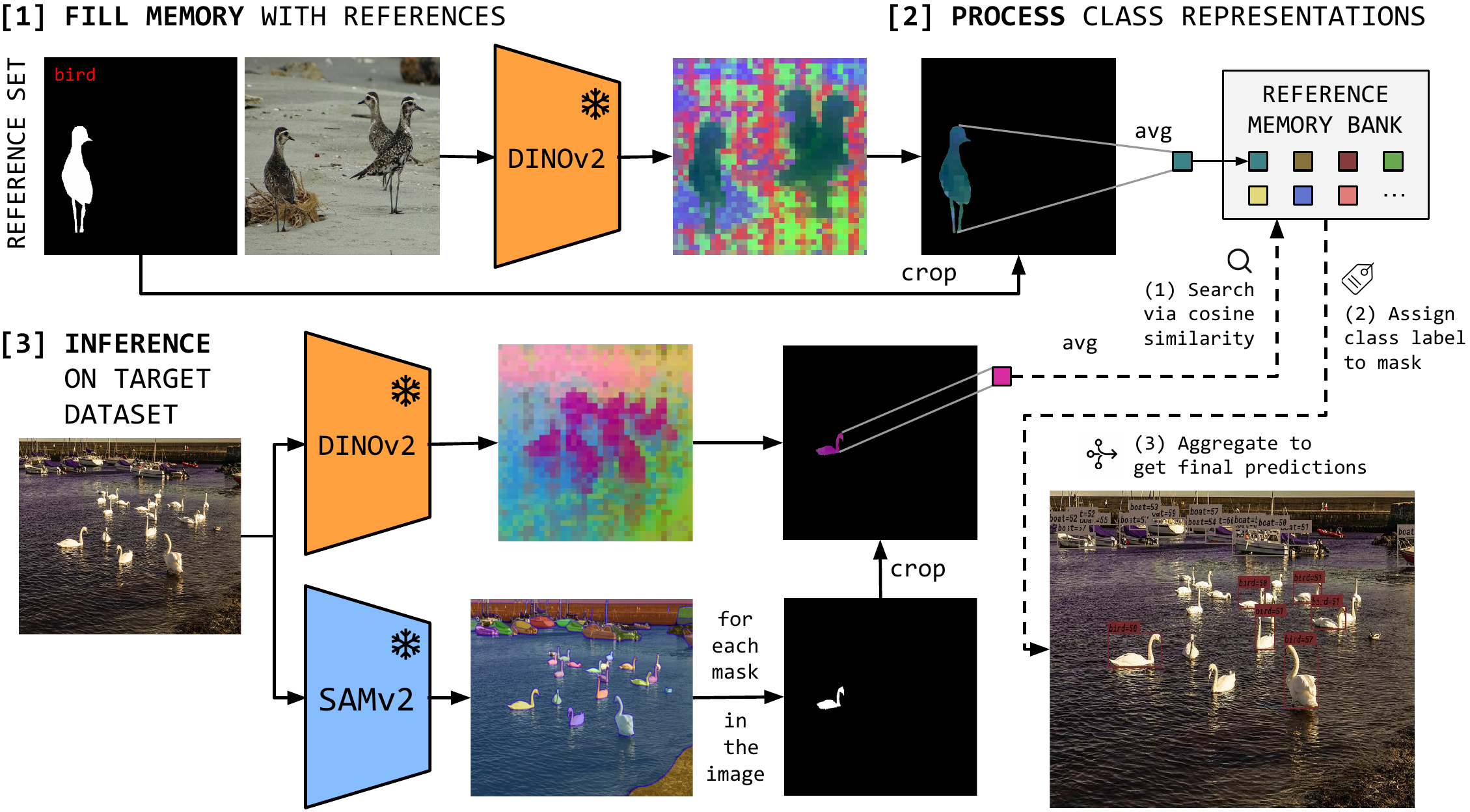}
  \caption{\textbf{Overview of our training-free method for few-shot instance segmentation and object detection.} (1) \textit{Reference Memory Creation}: A segmented reference image is processed using the DINOv2 model to generate semantic feature embeddings. (2) \textit{Feature aggregation}: We compute instance-wise feature representations, and then, aggregate them into class-wise prototypes, stored in the memory bank. (3) \textit{Inference on Target Dataset}: For each target image, SAMv2 generates instance segmentation masks while DINOv2 extracts semantic features. Using cosine similarity, each mask's embedding is compared with the reference memory bank to assign the most similar class label. Finally, predictions are aggregated via semantic-aware soft merging to produce the final annotated image. This pipeline enables semantic prompting via reference images, without requiring fine-tuning, and demonstrates state-of-the-art performance on established benchmarks (COCO-FSOD, PASCAL-FSOD) and strong generalization across domains (CD-FSOD).}
  \label{fig:architecture-overview}
\end{figure*}
}

\noindent\textbf{Automatic Vision Prompting for SAM} aims to construct automatic prompting pipelines to enhance SAM’s versatility in complex visual tasks, reducing its reliance on manual inputs. Training-free methods \cite{personaliseSAM, Liu2023May_matcher} leverage feature-matching techniques but often rely on manually tuned thresholds, distance metrics and complex pipelines. Other approaches focus on learning prompts directly, such as spatial or semantic optimisation \cite{visualprompttunning, sun2024vrp, Huang2024Jan_learning_to_prompt}, but these methods still face challenges in multi-instance or semantically dense settings. Zero-shot methods \cite{cpt, redcircle} introduce visual markers to guide attention during segmentation, but lack fine-grained precision. Recent efforts, including SEEM \cite{zou2023seem}, have unified segmentation and recognition tasks through shared decoders, while SINE \cite{sine} tackles task ambiguity by disentangling segmentation tasks. Others have paired SAM with Stable Diffusion for open-vocabulary segmentation \cite{diffews}, enabling it to incorporate semantic cues. Similarly, LISA \cite{lai2023lisa} utilises language-based instructions to adapt SAM for text-guided tasks. Despite these efforts, handling multi-instance scenarios and semantic ambiguity without training remains challenging. Our method differs by integrating SAM with DINOv2 without additional training or prompt optimisation. We introduce a multi-stage method (memory bank construction, representation aggregation, and semantic-aware feature matching) with fixed hyperparameters across all experimental settings, making it practical for a wide range of downstream applications and directly accessible to practitioners across 
domains.

\vspace{-2mm}
\section{Method}
\label{sec:method}
\subsection{Preliminaries}
\vspace{-1mm}
\textbf{Segment Anything Model (SAM)} \cite{sam} is designed for promptable segmentation, that is, it responds to different types of geometric prompts to generate
segmentation masks. It consists of three
components: an image encoder, a prompt encoder, and a mask decoder. The image encoder is a
Vision Transformer (ViT) \cite{vit} adapted for high-resolution inputs \cite{li2022exploring}. The prompt encoder handles
sparse (points, boxes, text) and dense (mask) prompts, which are encoded with positional encodings \cite{tancik2020fourier}, learnt embeddings, and off-the-shelf text encoders. The mask decoder
generates masks by using a modified Transformer decoder block \cite{detr} with self-attention and cross-attention on the prompts. For training, SAM uses a combination of focal loss \cite{focal_loss} and dice loss \cite{dice_loss}.

\textbf{DINOv2 vision encoder} \cite{dinov2} is a self-supervised ViT-based~\cite{vit} model
designed to produce general-purpose visual features using a discriminative self-supervised learning approach.
DINOv2 makes use of teacher-student networks, incorporates Sinkhorn-Knopp normalisation, multi-crops strategy, separate projection heads for image-level and patch-level objectives, and additional regularisation techniques to stabilise and scale training \cite{dino, ibot}. It is trained on a curated dataset (LVD-142M) of 142 million images using efficient training techniques
(memory-efficient attention, stochastic depth, etc.). Its features are transferable across tasks and domains, making it a robust backbone for
global and local visual understanding tasks.

\textbf{Reference-based instance segmentation} aims to segment a target image by using a reference segmented image
(rather than
geometric prompts).
Given a reference image $I_r$ and its
annotations $M^i_r$ (where $i$ denotes different object categories), we use this data to segment the corresponding regions in the target image $I_t$ that also belong to category $i$.

\subsection{Training-free method}

The goal
is to extract category-specific features from a set of annotated reference examples and use them to segment and classify instances in target images.
Unlike methods that require model retraining, we employ a memory-based approach to store discriminative representations for object categories. It consists of three main stages (see Figure~\ref{fig:architecture-overview}):
(1) \textit{constructing a memory bank}
from reference images,
(2) refining these representations via \textit{two-stage feature aggregation}, 
and (3) performing \textit{inference on the target images}
through feature matching and semantic-aware soft merging.

\textbf{(1) Memory Bank Construction.}
Given a set of reference images $\{ I_r^j \}_{j=1}^{N_r}$ and their 
instance masks $\{ M_r^{j,i} \}_{j=1}^{N_r}$ for category $i$, we extract dense feature maps $F_r^j \in \mathbb{R}^{H' \times W' \times d}$ using a
frozen encoder $\mathcal{E}$\footnote{{\scriptsize \url{https://github.com/facebookresearch/dinov2}}}, where $d$ is the feature dimension and $H', W'$ denote the spatial resolution of the feature map.
Instance masks $M_r^{j,i} \in \{0,1\}^{H' \times W'}$ are resized to match this resolution. For each category $i$, we store the masked features $\mathcal{F}_r^{j,i} = F_r^j \odot M_r^{j,i}$
where $\odot$ denotes element-wise multiplication. These category-wise feature sets are stored in a memory bank $\mathcal{M}_i$, which is synchronised across GPUs.

\textbf{(2) Two-stage feature aggregation.}
To construct category prototypes, we first compute instance-wise feature representations, and then, aggregate them into class-wise prototypes.

\renewcommand{\labelenumi}{(\alph{enumi})}
\begin{enumerate}
    
    \item \textbf{Instance-wise prototypes:} Each instance $k$ in reference image $I_r^j$ has its own prototype, computed by averaging the feature embeddings within its corresponding mask:
    $
    P_r^{j,k} = \frac{1}{\| M_r^{j,k} \|_1} \sum_{(u,v)} M_r^{j,k}(u,v) F_r^j(u,v)
    $
    where $P_r^{j,k} \in \mathbb{R}^d$ is the mean feature representation of the $k$-th instance in image $I_r^j$.

    \item \textbf{Class-wise prototype:} We compute the category prototype $P_i$ by averaging all instance-wise prototypes belonging to the same category $i$: $P_i = \frac{1}{N_i} \sum_{j=1}^{N_r} \sum_{k \in \mathcal{K}_i^j} P_r^{j,k},$
    where $\mathcal{K}_i^j$ is the set of instances in image $I_r^j$ that belong to category $i$, and $N_i = \sum_{j=1}^{N_r} |\mathcal{K}_i^j|$ is the total number of instances belonging to category $i$. These class-wise prototypes $P_i$ are stored in the memory bank.

\end{enumerate}

\textbf{(3) Inference on Target Images.}
Given a target image $I_t$, we extract dense features $F_t \in \mathbb{R}^{H' \times W' \times d}$ using 
$\mathcal{E}$. We use
SAM
to generate $N_m$ candidate instance masks $\{ M_t^m \}_{m=1}^{N_m}$, where $M_t^m \in \{0,1\}^{H' \times W'}$. Each mask $M_t^m$ is used to compute a feature representation $P_t^m$ via average pooling and L2-norm:
$P_t^m = \frac{1}{\| M_t^m \|_1} \sum_{(u,v)} M_t^m(u,v) F_t(u,v),$ $\quad \hat{P}_t^m = \frac{P_t^m}{\| P_t^m \|_2}$ where $ \hat{P}_t^m \in \mathbb{R}^d$.

To classify each candidate mask, we perform:

\begin{enumerate}

    \item \textbf{Feature Matching.} We compute the cosine similarity between $\hat{P}_t^m$ and category prototypes $P_i$, which provides the classification score $S_t^m$ for mask $M_t^m$: $
    S_t^m = \max_i \left( \frac{\hat{P}_t^m \cdot P_i}{\| P_i \|_2} \right)$

    \item \textbf{Semantic-Aware Soft Merging.}
    To handle overlapping predictions, we introduce a novel soft merging strategy. Given two masks $M_t^m$ and $M_t^{m'}$ of the same category, we compute their intersection-over-self (IoS), and weight it by feature similarity.
    
    \vspace{-4mm}
    { 
    \small \[
    \text{IoS}(M_t^m, M_t^{m'})\!=\!\frac{\sum (M_t^m \cap M_t^{m'})}{\sum M_t^m}
    \quad
    w_{m,m'}\!=\!\frac{\hat{P}_t^m \cdot \hat{P}_t^{m'}}{\|\hat{P}_t^{m'}\|_2}
    \]
    }
    
    The final score for each mask is adjusted using a decay factor
    $
    S_t^m\!\leftarrow\!S_t^m\!\cdot\!\sqrt{(1\!-\!\text{IoS}(M_t^m,\,M_t^{m'}) w_{m,m'}})
    $,
    reducing redundant detections while preserving distinct instances that may partially overlap. We rank masks by their adjusted scores, and select top-$K$ predictions.

\end{enumerate}

\subsection{Technical implementation details}

Our code (available in Supp.) builds upon SAM2-L (Hierarchical ViT) for mask generation and DINOv2-L as 
feature encoder. The encoder processes images at 518×518 resolution with a patch size of 14×14, while SAM2 operates at 1024×1024 resolution. During inference, SAM2
generates candidate masks using a 32×32 grid of query points. For each mask, we compute L2-normalised features by average pooling the encoder features within the masked region. These features are compared against our memory bank, which stores features from $n$ reference images per category. We employ non-maximum suppression with an IoU threshold of 0.5, followed by our semantic-aware soft merging strategy to handle overlapping predictions. The model outputs up to 100 instances per image. The  implementation uses PyTorch \cite{Ansel_PyTorch_2_Faster_2024} and PyTorch Lightning \cite{pytorchlightning} for distributed workload across GPUs.

\section{Results}
\label{sec:results}

\subsection{Object Detection and Instance Segmentation}

Although our training-free method outputs segmentation masks, we convert instance masks to bounding boxes for a fair comparison with existing methods.

\textbf{COCO-FSOD Benchmark.} We evaluate our method in a strict few-shot setting on the COCO-20$^i$ dataset \cite{coco, coco-novel}, using the standard 10-shot and 30-shot settings. Results for the COCO-FSOD benchmark are shown in Table \ref{tab:coco-fsod}. All results are reported for COCO-NOVEL classes. Novel classes are the COCO categories that intersect with PASCAL VOC categories \cite{pascal}. Our method achieves state-of-the-art while being completely training-free, outperforming approaches that fine-tune on novel classes. Figure \ref{fig:coco-grid} presents qualitative results, showing our method's ability to handle multiple overlapping instances in crowded scenes with fine-grained semantics and precise localisation. With semantic-aware soft merging, we mitigate duplicate detections and false positives. Failure cases are discussed in the Appendix~\ref{app:failure-cases}.

{\begin{table}[h]
\centering
\begin{adjustbox}{width=\linewidth}
\addtolength{\tabcolsep}{-0.45em}
\begin{tabular}{lccccccc}
\toprule
\multicolumn{1}{c}{\multirow{2}{*}{\textbf{Method}}} & \multirow{2}{*}{\makecell{\textbf{FT}}}  & \multicolumn{3}{c}{\textbf{10-shot}} & \multicolumn{3}{c}{\textbf{30-shot}} \\
\cmidrule(lr){3-5} \cmidrule(lr){6-8}
& & \textbf{\small nAP} & \textbf{\small nAP50} & \textbf{\small nAP75} & \textbf{\small nAP} & \textbf{\small nAP50} & \textbf{\small nAP75} \\
\midrule
TFA{\small~\cite{tfa}}                             & \cmark & 10.0 & 19.2 & 9.2 & 13.5 & 24.9 & 13.2 \\
FSCE{\small~\cite{fsce}}                           & \cmark & 11.9 & -- & 10.5 & 16.4 & -- & 16.2 \\
Retentive RCNN{\small~\cite{retentive-rcnn}}       & \cmark & 10.5 & 19.5 & 9.3 & 13.8 & 22.9 & 13.8 \\
HeteroGraph{\small~\cite{heterograph}}             & \cmark & 11.6 & 23.9 & 9.8 & 16.5 & 31.9 & 15.5 \\
Meta F. R-CNN{\small~\cite{meta-faster-rcnn}}      & \cmark & 12.7 & 25.7 & 10.8 & 16.6 & 31.8 & 15.8 \\
LVC~\cite{lvc}                             & \cmark & 19.0 & 34.1 & 19.0 & 26.8 & 45.8 & 27.5 \\
C. Transformer{\small~\cite{cross-transformer}}    & \cmark & 17.1 & 30.2 & 17.0 & 21.4 & 35.5 & 22.1 \\
NIFF{\small~\cite{niff}}                           & \cmark & 18.8 & -- & -- & 20.9 & -- & -- \\
DiGeo{\small~\cite{digeo}}                         & \cmark & 10.3 & 18.7 & 9.9 & 14.2 & 26.2 & 14.8 \\
CD-ViTO {\small(ViT-L) \cite{cd-fsod}}    & \cmark & 35.3 & 54.9 & 37.2 & 35.9 & 54.5 & 38.0 \\
\midrule
FSRW{\small~\cite{fsrw}}                           & \xmark & 5.6 & 12.3 & 4.6 & 9.1 & 19.0 & 7.6 \\
Meta R-CNN{\small~\cite{metarcnn}}                 & \xmark & 6.1 & 19.1 & 6.6 & 9.9 & 25.3 & 10.8 \\
DE-ViT (L) {\small\cite{devit}}                & \xmark & 34.0 & 53.0 & 37.0 & 34.0 & 52.9 & 37.2 \\
\textbf{Training-free (ours)}              & \xmark & \textbf{36.6} & \textbf{54.1} & \textbf{38.3} & \textbf{36.8} & \textbf{54.5} & \textbf{38.7} \\
\bottomrule
\end{tabular}
\end{adjustbox}
\caption{Comparison of our training-free method against state-of-the-art approaches on the COCO-FSOD (10-shot and 30-shot). We achieve SOTA performance without finetuning on novel classes. Results are reported as nAP, nAP50, and nAP75. nAP refers to mAP for novel classes. Competing
results are sourced from 
\cite{cd-fsod}. Since we are the only method
providing
both bounding box and segmentation results,
we omit segmentation AP on this table. FT refers to finetuning on novel classes.
}
\label{tab:coco-fsod}
\vspace{-2mm}
\end{table}
}

{\begin{figure}[t]
  \centering
  \includegraphics[width=0.97\linewidth]{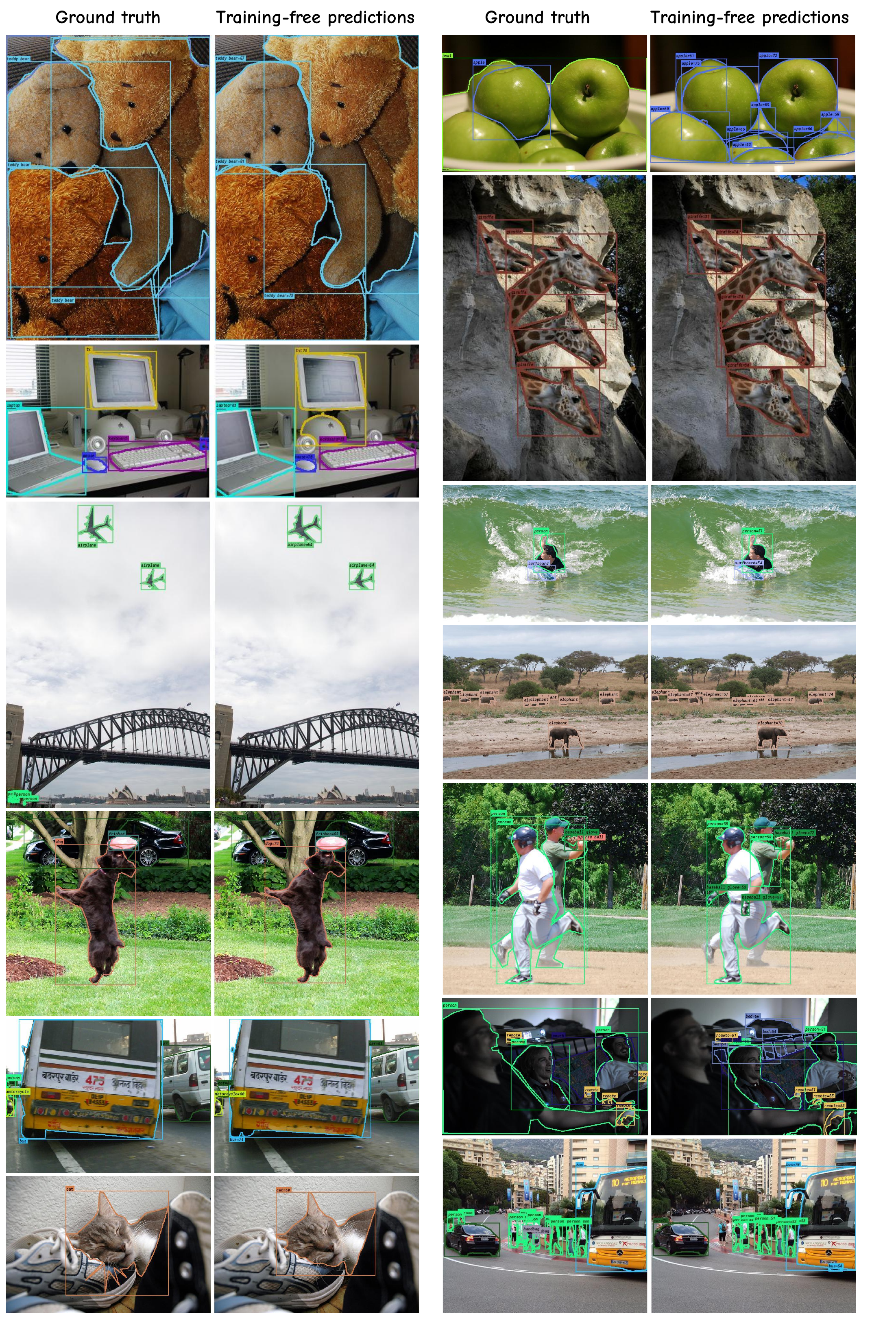}
  \caption{
  Qualitative results on the COCO val2017 test set under the 10-shot setting (using 10 reference images per class). Bounding box visualisations are thresholded at 0.5. Our method effectively handles multiple overlapping instances in crowded scenes, demonstrating fine-grained semantics and precise localisation. Through semantic-aware soft merging, we avoid duplicate detections and false positives. Best viewed when zoomed in.
  }
  \label{fig:coco-grid}
  \vspace{-2mm}
\end{figure}
}

\textbf{PASCAL VOC Few-Shot Benchmark.} The PASCAL-VOC dataset \cite{pascal} consists of 20 classes. For few-shot evaluation, we adopt the standard approach \cite{niff}, splitting the classes into three groups, each with 15 base and 5 novel classes. As in prior work \cite{devit}, we report AP50 results on the novel classes. Table \ref{tab:pascal-fsod} shows that our method outperforms all previous approaches across all splits, achieving state-of-the-art performance across all splits. This holds for
methods that fine-tune on novel classes and those that do not.

{\begin{table*}[t]
\centering
\resizebox{\textwidth}{!}{
\addtolength{\tabcolsep}{-0.25em}
\begin{tabular}{lccccccccccccccccc}
\toprule
\multicolumn{1}{c}{\multirow{2}{*}{\textbf{Method}}} & \multirow{2}{*}{\makecell{\textbf{FT}}} & \multicolumn{5}{c}{\textbf{Novel Split 1}} & \multicolumn{5}{c}{\textbf{Novel Split 2}} & \multicolumn{5}{c}{\textbf{Novel Split 3}} & \multirow{2}{*}{\textbf{Avg}} \\
\cmidrule(lr){3-7} \cmidrule(lr){8-12} \cmidrule(lr){13-17}
 & & \textbf{1} & \textbf{2} & \textbf{3} & \textbf{5} & \textbf{10} & \textbf{1} & \textbf{2} & \textbf{3} & \textbf{5} & \textbf{10} & \textbf{1} & \textbf{2} & \textbf{3} & \textbf{5} & \textbf{10} &  \\
\midrule
FsDetView \cite{fsdetview}                  & \cmark & 25.4 & 20.4 & 37.4 & 36.1 & 42.3 & 22.9 & 21.7 & 22.6 & 25.6 & 29.2 & 32.4 & 19.0 & 29.8 & 33.2 & 39.8 & 29.2 \\
TFA \cite{tfa}                              & \cmark & 39.8 & 36.1 & 44.7 & 55.7 & 56.0 & 23.5 & 26.9 & 34.1 & 35.1 & 39.1 & 30.8 & 34.8 & 42.8 & 49.5 & 49.8 & 39.9 \\
Retentive RCNN \cite{retentive-rcnn}        & \cmark & 42.4 & 45.8 & 45.9 & 53.7 & 56.1 & 21.7 & 27.8 & 35.2 & 37.0 & 40.3 & 30.2 & 37.6 & 43.0 & 49.7 & 50.1 & 41.1 \\
DiGeo \cite{digeo}                          & \cmark & 37.9 & 39.4 & 48.5 & 58.6 & 61.5 & 26.6 & 28.9 & 41.9 & 42.1 & 49.1 & 30.4 & 40.1 & 46.9 & 52.7 & 54.7 & 44.0 \\
HeteroGraph \cite{heterograph}              & \cmark & 42.4 & 51.9 & 55.7 & 62.6 & 63.4 & 25.9 & 37.8 & 46.6 & 48.9 & 51.1 & 35.2 & 42.9 & 47.8 & 54.8 & 53.5 & 48.0 \\
Meta Faster R-CNN \cite{meta-faster-rcnn}   & \cmark & 43.0 & 54.5 & 60.6 & 66.1 & 65.4 & 27.7 & 35.5 & 46.1 & 47.8 & 51.4 & 40.6 & 46.4 & 53.4 & 59.9 & 58.6 & 50.5 \\
CrossTransformer \cite{cross-transformer}   & \cmark & 49.9 & 57.1 & 57.9 & 63.2 & 67.1 & 27.6 & 34.5 & 43.7 & 49.2 & 51.2 & 39.5 & 54.7 & 52.3 & 57.0 & 58.7 & 50.9 \\
LVC \cite{lvc}                              & \cmark & 54.5 & 53.2 & 58.8 & 63.2 & 65.7 & 32.8 & 29.2 & 50.7 & 49.8 & 50.6 & 48.4 & 52.7 & 55.0 & 59.6 & 59.6 & 52.3 \\
NIFF \cite{niff} (*)                        & \cmark & 62.8 & 67.2 & 68.0 & 70.3 & 68.8 & 38.4 & 42.9 & 54.0 & 56.4 & 54.0 & 56.4 & 62.1 & 61.2 & 64.1 & 63.9 & 59.4 \\
\midrule
Multi-Relation Det \cite{multirelation-det} & \xmark & 37.8 & 43.6 & 51.6 & 56.5 & 58.6 & 22.5 & 30.6 & 40.7 & 43.1 & 47.6 & 31.0 & 37.9 & 43.7 & 51.3 & 49.8 & 43.1 \\
DE-ViT (ViT-S/14) \cite{devit}              & \xmark & 47.5 & 64.5 & 57.0 & 68.5 & 67.3 & 43.1 & 34.1 & 49.7 & 56.7 & 60.8 & 52.5 & 62.1 & 60.7 & 61.4 & 64.5 & 56.7 \\
DE-ViT (ViT-B/14) \cite{devit}              & \xmark & 56.9 & 61.8 & 68.0 & 73.9 & 72.8 & 45.3 & 47.3 & 58.2 & 59.8 & 60.6 & 58.6 & 62.3 & 62.7 & 64.6 & 67.8 & 61.4 \\
DE-ViT (ViT-L/14) \cite{devit}              & \xmark & 55.4 & 56.1 & 68.1 & 70.9 & 71.9 & 43.0 & 39.3 & 58.1 & 61.6 & 63.1 & 58.2 & 64.0 & 61.3 & 64.2 & 67.3 & 60.2 \\

\textbf{Training-free (ours)} & \xmark & \textbf{70.8} & \textbf{72.3} & \textbf{73.3} & \textbf{77.2} & \textbf{79.1} & \textbf{54.5} & \textbf{67.0} & \textbf{76.3} & \textbf{75.9} & \textbf{78.2} & \textbf{61.1} & \textbf{67.9} & \textbf{71.3} & \textbf{70.8} & \textbf{72.6} & \textbf{71.2} \\
\bottomrule
\end{tabular}%
}
\caption{
AP50 results on the novel classes of the Pascal VOC few-shot benchmark. Competing method results are sourced from \cite{devit}. State-of-the-art results are highlighted in \textbf{bold}. (*) indicates that the corresponding implementation is not publicly accessible. Our proposed training-free approach consistently achieves superior performance across all splits, outperforming fine-tuned methods. FT refers to finetuning on novel classes.
}

\label{tab:pascal-fsod}
\vspace{-1mm}
\end{table*}
}

\subsection{Cross-Domain Few-Shot Object Detection}

{\begin{table*}[htbp]
\centering
\begingroup
\footnotesize
\setlength{\tabcolsep}{2.5pt}
\renewcommand{\arraystretch}{1.05}
\begin{adjustbox}{max width=\linewidth}
\begin{tabular}{lcccccccc}
\toprule
\textbf{Method} & \textbf{FT} & \textbf{\makecell{ArT\\axOr}} & \textbf{\makecell{Clip\\art1k}} & \textbf{DIOR} & \textbf{\makecell{Deep\\Fish}} & \textbf{\makecell{NEU\\DET}} & \textbf{\makecell{UO\\DD}} & \textbf{Avg} \\
\toprule
\multicolumn{9}{c}{\textbf{Fine-tuned on novel classes}} \\
\midrule
TFA w/cos $\circ$ \cite{tfa}              & \cmark & 3.1/8.8/14.8 & -/-/- & 8.0/18.1/20.5 & -/-/- & -/-/- & 4.4/8.7/11.8 & -/-/- \\
FSCE $\circ$ \cite{fsce}                  & \cmark & 3.7/10.2/15.9 & -/-/- & 8.6/18.7/21.9 & -/-/- & -/-/- & 3.9/9.6/12.0 & -/-/- \\
DeFRCN $\circ$ \cite{defrcn}              & \cmark & 3.6/9.9/15.5 & -/-/- & 9.3/18.9/22.9 & -/-/- & -/-/- & 4.5/9.9/12.1 & -/-/- \\
Distill-cdfsd $\circ$ \cite{cdfsod-bench} & \cmark & 5.1/12.5/18.1 & 7.6/23.3/27.3 & 10.5/19.1/26.5 & nan/15.5/15.5 & nan/16.0/21.1 & 5.9/12.2/14.5 & -/16.4/20.5 \\
ViTDeT-FT$\dag$ \cite{vitdetft}           & \cmark & 5.9/20.9/23.4 & 6.1/23.3/25.6 & 12.9/23.3/29.4 & 0.9/9.0/6.5 & 2.4/13.5/15.8 & 4.0/11.1/15.6 & 5.4/16.9/19.4 \\
Detic-FT$\dag$ \cite{detic}               & \cmark & 3.2/8.7/12.0 & 15.1/20.2/22.3 & 4.1/12.1/15.4 & 9.0/14.3/17.9 & 3.8/14.1/16.8 & 4.2/10.4/14.4 & 6.6/13.3/16.5 \\
DE-ViT-FT$\dag$ \cite{devit}              & \cmark & 10.5/38.0/49.2 & 13.0/38.1/40.8 & 14.7/23.4/25.6 & 19.3/21.2/21.3 & 0.6/7.8/8.8 & 2.4/5.0/5.4 & 10.1/22.3/25.2 \\
CD-ViTO$\dag$ \cite{cd-fsod}              & \cmark & 21.0/47.9/60.5 & 17.7/41.1/44.3 & 17.8/26.9/30.8 & 20.3/22.3/22.3 & 3.6/11.4/12.8 & 3.1/6.8/7.0 & 13.9/26.1/29.6 \\
\midrule
\multicolumn{9}{c}{\textbf{Training-free (no novel class fine-tuning)}} \\
\midrule
Meta-RCNN $\circ$ \cite{metarcnn}         & \xmark & 2.8/8.5/14.0 & -/-/- & 7.8/17.7/\textbf{20.6} & -/-/- & -/-/- & 3.6/8.8/11.2 & -/-/- \\
Detic$\dag$ \cite{detic}                  & \xmark & 0.6/0.6/0.6 & 11.4/11.4/11.4 & 0.1/0.1/0.1 & 0.9/0.9/0.9 & 0.0/0.0/0.0 & 0.0/0.0/0.0 & 2.2/2.2/2.2 \\
DE-ViT$\dag$ \cite{devit}                 & \xmark & 0.4/10.1/9.2 & 0.5/5.5/11.0 & 2.7/7.8/8.4 & 0.4/2.5/2.1 & 0.4/1.5/1.8 & 1.5/3.1/3.1 & 1.0/5.1/5.9 \\
\textbf{Training-free (ours)}             & \xmark & \textbf{28.2/35.7/35.0} & \textbf{18.9/24.9/25.9} & \textbf{14.9/18.5}/16.4 & \textbf{30.5/29.6/29.6} & \textbf{5.5/5.2/5.5} & \textbf{10.0/20.2/16.0} & \textbf{18.0/22.4/21.4} \\
\bottomrule
\end{tabular}
\end{adjustbox}
\caption{Per-cell values are 1-shot/5-shot/10-shot mAP on the CD-FSOD benchmark. The $\circ$ symbol indicates results sourced from Distill-cdfsod \citep{cdfsod-bench}, while $\dag$ denotes results reported by CD-ViTO \citep{cd-fsod}. `Avg.' represents the average performance across datasets. FT refers to finetuning on novel classes.}
\label{tab:cdfsod}
\endgroup
\vspace{-1mm}
\end{table*}

}

The CD-FSOD benchmark \citep{cd-fsod} evaluates
cross-domain few-shot object detection 
models by addressing challenges in domain shifts and limited data scenarios. It uses COCO as the source training dataset (SD), and six target datasets (TD) — ArTaxOr, Clipart1k, DIOR, DeepFish, NEUDET, and UODD — spanning photorealistic, cartoon, aerial, underwater, and industrial domains with high inter-class variance. While many approaches fine-tune on
labeled instances (support set S) from TD before testing on the query set Q, our model is
\textit{training-free}. Thus, we directly evaluate on the six target datasets without
fine-tuning.
Table \ref{tab:cdfsod} compares FSOD methods on CD-FSOD
across 1, 5, and 10-shot settings. Our method
is 
state-of-the-art among training-free approaches and remains competitive with fine-tuned models. These results demonstrate its strong cross-domain generalisation and robustness without
retraining.

{\begin{figure*}[!htbp]
\centering
\begin{minipage}{0.35\linewidth}
  \includegraphics[width=\linewidth]{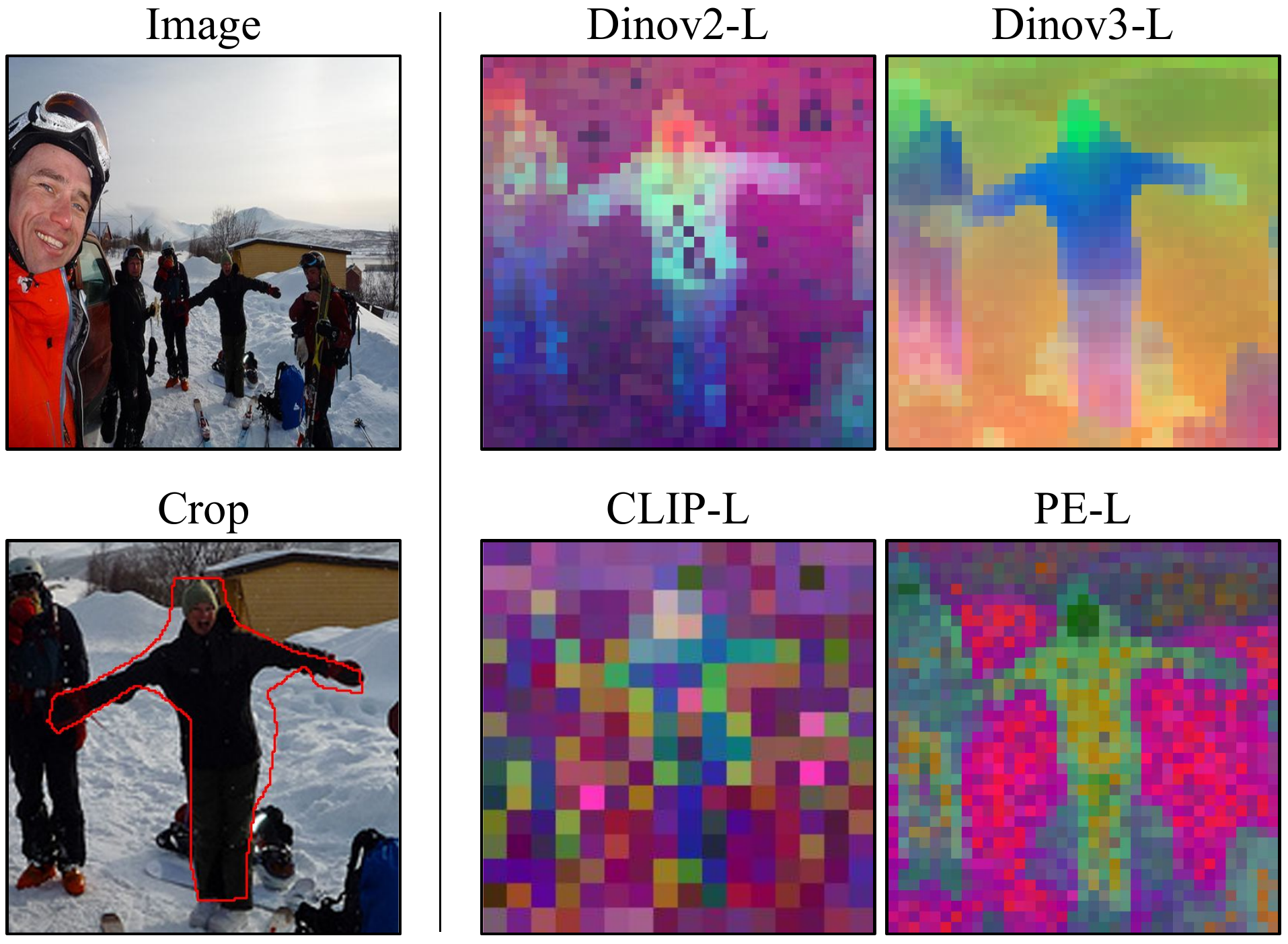}
  \caption{
  \textit{Feature comparison across semantic backbones}. Left: full image and cropped region of interest. Right: feature PCA
  maps across backbones.
  CLIP features are low-resolution and irregular, PE-Spatial
  is noisy but informative, and DINOv2/DINOv3
  are spatially consistent and
  structured.
  More visualisations in Figure~\ref{fig:feature-comparison-large}.
  }
  \label{fig:feature-comparison-small}
\end{minipage}
\hfill
\begin{minipage}{0.63\linewidth}
  \centering
  \includegraphics[width=\linewidth]{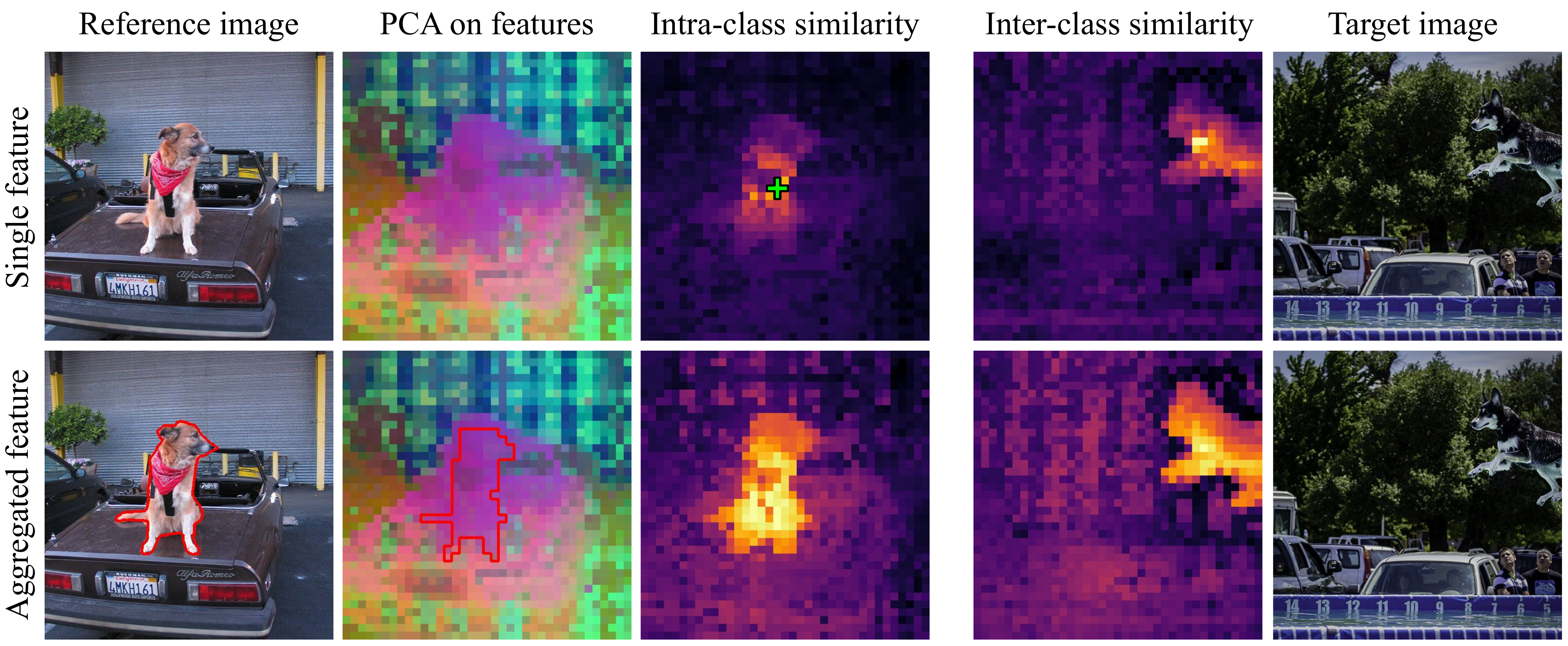}
  \caption{
  \textit{Single vs. aggregated feature similarity.} We compare cosine similarity maps obtained from (top) a single DINOv2 patch feature selected inside the reference mask (marked with {\contour{black}{\textcolor{green}{\textbf{+}}}}) and (bottom) the aggregated prototype obtained by averaging all features within the mask. For each case, we show intra-class similarity (within the same image) and inter-class similarity (with a target image). Single-feature similarity highlights only local object parts, whereas aggregated features produce more coherent, object-level similarity patterns.}
  \label{fig:feature-similarity-small}
\end{minipage}
\vspace{-6mm}
\end{figure*}
}

{\begin{figure*}[t]
  \centering
  \includegraphics[width=0.9\linewidth]{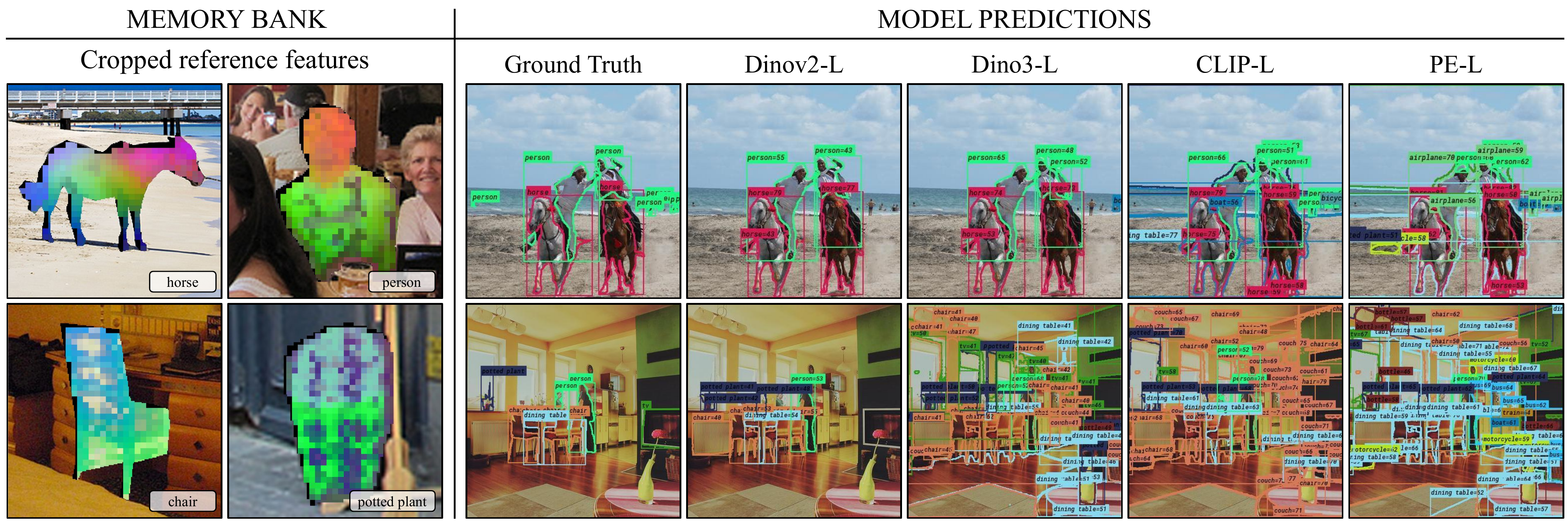}
  \caption{
  \textit{Memory-bank visualisation and model predictions across backbones}. (Left) Memory-bank: reference images overlaid with their masked features. (Right) Model predictions: outputs from our pipeline with different backbones (DINOv2, DINOv3, CLIP, and PE-Spatial) along with ground truth masks for comparison. These examples illustrate how differences in feature quality influence downstream detection and segmentation. Additional outputs are provided in Figure \ref{fig:output-comparison-large}.}
  \label{fig:output-comparison-small}
  \vspace{-2mm}
\end{figure*}
}

\subsection{Ablations}

\textbf{Model Backbone.}
To evaluate the method transferability across foundation models, we replace
the semantic encoder DINOv2 with CLIP~\cite{clip}, DINOv3~\cite{dinov3}, and PE-Spatial~\cite{perception_encoder}; and the SAM-based segmenter with SAM variants.
Results are reported in Table \ref{tab:ablation-backbones}. Our pipeline transfers across backbones with moderate performance variation. CLIP models exhibit a performance drop, due to their low-resolution and noisier feature maps. In contrast, PE-Spatial and DINOv3 achieve competitive results with
modest degradation. All models are used off-the-shelf without hyperparameter adjustments. 

Figure \ref{fig:feature-comparison-small} visualises backbones feature maps with PCA. CLIP features show low spatial detail and irregular patterns, while PE-Spatial features are noisy but discriminative. DINOv2 and DINOv3 produce consistent, high-resolution features, aligning with their stronger performance.
Figure \ref{fig:output-comparison-small} illustrates the
predictions across backbones, alongside memory-bank exemplars with their
PCA feature visualisations. More examples
are provided in Figure \ref{fig:feature-comparison-large} and Figure \ref{fig:output-comparison-large} respectively. These results show that it is not tied to specific backbone and transfers across a range of foundation encoders.

\textbf{Feature matching.} We compare patch-level and prototype-level feature matching using DINOv2 (Fig.~\ref{fig:feature-similarity-small}). While single patch features capture local appearance, they lack full object coverage and yield unstable similarity responses. Averaging features over the reference mask produces object-aligned prototypes, resulting in more spatially coherent and stable similarity maps for memory-bank matching. Additional examples are shown in Figures~\ref{fig:feature-similarity-large} and \ref{fig:feature-similarity-large-extras}.

\begin{table}[!h]
\centering
\begin{adjustbox}{width=0.95\linewidth}
\addtolength{\tabcolsep}{-0.4em}
\begin{tabular}{lccccc}
\toprule
\multicolumn{1}{c}{\multirow{3}{*}{\makecell{\textbf{Semantic} \\ \textbf{backbone}}}} & \multicolumn{1}{c}{\multirow{3}{*}{\makecell{\textbf{Segm.} \\ \textbf{backb.}}}} & \multicolumn{2}{c}{\textbf{10-shot}} & \multicolumn{2}{c}{\textbf{30-shot}} \\
\cmidrule(lr){3-4} \cmidrule(lr){5-6}
& & \multicolumn{1}{c}{\textbf{bbox}} & \multicolumn{1}{c}{\textbf{segm}} & \multicolumn{1}{c}{\textbf{bbox}} & \multicolumn{1}{c}{\textbf{segm}} \\
\cmidrule(lr){3-3} \cmidrule(lr){4-4} \cmidrule(lr){5-5} \cmidrule(lr){6-6}
& & \textbf{\small nAP} & \textbf{\small nAP} & \textbf{\small nAP} & \textbf{\small nAP} \\
\midrule
DINOv2-ViT-L-14 & SAM2-L
    & 35.7 & 33.3 & 36.8 & 34.2 \\
\midrule
DINOv3-ViT-B-16 & SAM2-L
    & 32.8 & 30.9 & 33.5 & 31.6 \\
DINOv3-ViT-L-16 & SAM2-L
    & 33.8 & 32.2 & 34.4 & 32.8 \\
DINOv3-ViT-H-16+ & SAM2-L
    & 25.5 & 24.4 & 26.2 & 25.0 \\ %
DINOv3-ViT-H-16+ (px) & SAM2-L
    & 26.8 & 25.5 & 27.6 & 26.3 \\
CLIP-ViT-B-32 & SAM2-L
    & 15.2 & 13.9 & 15.8 & 14.6 \\
CLIP-ViT-B-16 & SAM2-L
    & 19.2 & 18.1 & 19.5 & 18.3 \\
CLIP-ViT-L-14 & SAM2-L
    & 18.6 & 17.3 & 18.7 & 17.4 \\
CLIP-ViT-L-14-336px & SAM2-L
    & 17.8 & 16.7 & 17.7 & 16.7 \\
PE-Spatial-L-14-448 (PE) & SAM2-L
    & 25.7 & 24.1 & 26.5 & 24.7 \\
PE-Spatial-L-14-448 (IN) & SAM2-L
    & 26.5 & 24.7 & 27.3 & 25.4 \\
PE-Spatial-G-14-448 (PE) & SAM2-L
    & 24.7 & 23.1 & 24.9 & 23.3 \\
\midrule
DINOv2-ViT-L-14 & SAM2-T
    & 27.0 & 26.1 & 27.9 & 26.9 \\
DINOv2-ViT-L-14 & SAM2-S
    & 29.8 & 27.7 & 30.6 & 28.5 \\
DINOv2-ViT-L-14 & SAM2-B+
    & 29.6 & 28.4 & 30.4 & 29.2 \\
\bottomrule
\end{tabular}

\end{adjustbox}
\caption{
Backbone ablation on COCO-FSOD. We compare different semantic encoders (DINOv2, DINOv3, CLIP, PE-Spatial) paired with the same SAM-based (hiera) segmenter. Similarly, we compare different SAM model sizes paired with the same semantic backbone. Results are reported for 10-shot and 30-shot settings using nAP, nAP50, and nAP75. All experiments use a single seed (33) and no hyperparameter tuning (default prompting and aggregation settings). Results show that the pipeline remains functional across encoder choices, with performance largely influenced by feature-map resolution. PE refers to model's default normalisation, and IN indicates ImageNet normalisation. Model sizes are tiny (T), base (B), large (L), huge (H). Image pixels (px) indicates a larger input image size. 14, 16, 32 refer to the model's patch size.
}
\label{tab:ablation-backbones}
\vspace{-6mm}
\end{table}

\textbf{Do VLMs provide a competitive alternative?}
We evaluate a Qwen2.5-VL-7B + SAM2 pipeline on COCO few-shot and observe lower detection and segmentation performance. Details are in Appendix~\ref{app:vlm}.

\textbf{Do reference images impact performance?}
Reference images affect performance in low-shot settings, with variance decreasing as the number of references grows. Analysis of COCO annotations shows that larger, more centered masks yield stronger reference prototypes, motivating simple selection heuristics that improve one-shot performance without adding more references. Our method is robust to substantial reference degradation (e.g., blur), indicating limited sensitivity to reference quality. Full analysis in Appendix~\ref{app:reference-image}.

\textbf{Pipeline design.}
We analise pipeline components in Appendix~\ref{app:pipeline-design}.

\textbf{Why do similar classes remain challenging across backbones?}
Semantically similar categories exhibit overlapping feature embeddings, indicating that errors stem from backbone feature geometry rather than prototype construction. DINOv3 shows similar entanglement and produces smoother features that further reduce discriminability during prototype matching (Appendix~\ref{app:feature-quality-separability}).

\textbf{Runtime and memory efficiency.}
We report inference speed, memory usage, and model size, showing that the training-free pipeline remains efficient with cached references and lightweight matching (Appendix~\ref{app:implementation}).

\section{Conclusion}
\label{sec:conclusions}

In this work, we introduce a novel training-free approach for few-shot instance segmentation by integrating SAM's mask generation capabilities with the fine-grained semantic understanding of DINOv2. Our method uses reference images to construct a memory bank, refines its internal representations with feature aggregation, and performs feature matching for novel instances using cosine similarity and semantic-aware soft merging. We demonstrate that careful engineering of existing frozen foundation models can lead to state-of-the-art performance without the need for additional training: we achieve 36.8\% nAP on COCO-FSOD (outperforming fine-tuned methods), 71.2\% nAP50 on PASCAL VOC Few-Shot, and strong generalisation across domains (CD-FSOD benchmark). Furthermore, our semantic segmentation results show that our approach can be extended beyond instance segmentation by aggregating instance predictions into semantic maps.

For future work, we identify several promising directions: (1) exploring learning-based strategies to automatically find the most informative reference images for 1-5 shot scenarios; (2) addressing DINOv2's global semantic biases by improving feature localisation, especially for fine-grained tasks; and (3) investigating lightweight finetuning approaches to improve the internal memory bank representations for 1-5 shot scenarios.

\subsubsection*{LLM Usage}
\vspace{-1mm}
Within the scope of this paper, LLMs are used only to aid and polish writing.

\section*{Acknowledgements}

Funding for this research is provided in part by the SENSE~-~Centre for Satellite Data in Environmental Science CDT studentship (NE/T00939X/1), and an EPSRC New Investigator Award (EP/X020703/1). This work used JASMIN, the UK’s collaborative data analysis environment~\url{https://jasmin.ac.uk} \cite{jasmin}.

\section*{Impact Statement}

This work aims to advance the field of machine learning. While it may have broader societal implications, we do not identify any specific impacts that require explicit discussion.

\bibliography{references}
\bibliographystyle{icml2026}

\newpage
\appendix

{\newpage

\section{Appendix}

\subsection{Reference-Image Ablation}
\label{app:reference-image}

To understand why different reference images lead to performance variation, we observe that randomly sampled reference sets (via different seeds) yield different performances, particularly in low-shot settings. We analyse the statistical properties of COCO novel classes annotations, evaluate per-class reference-image sensitivity, and design simple heuristics for selecting higher-scoring references.

\subsubsection{Variance in reference set.}

Using different randomly sampled reference images leads to variations in results, as performance depends on the quality of the selected reference images for each class.
To quantify this variance, we evaluate our method on the COCO-20$^i$ few-shot object detection benchmark using different random seeds to select reference images. Figure \ref{fig:ablation-std} shows the standard deviation (std) across 10 runs. We observe that increasing the number of reference images (higher n-shots) reduces result variance, with lower std values. In 1, 2, and 3-shot settings, reference image selection has a more noticeable impact, while for 5 or more shots, the low std demonstrates robustness to reference set variation. This suggests that particularly in low-shot settings, performance is sensitive to the choice of individual reference images.

{\begin{figure}[h]
  \centering
  \includegraphics[width=\linewidth]{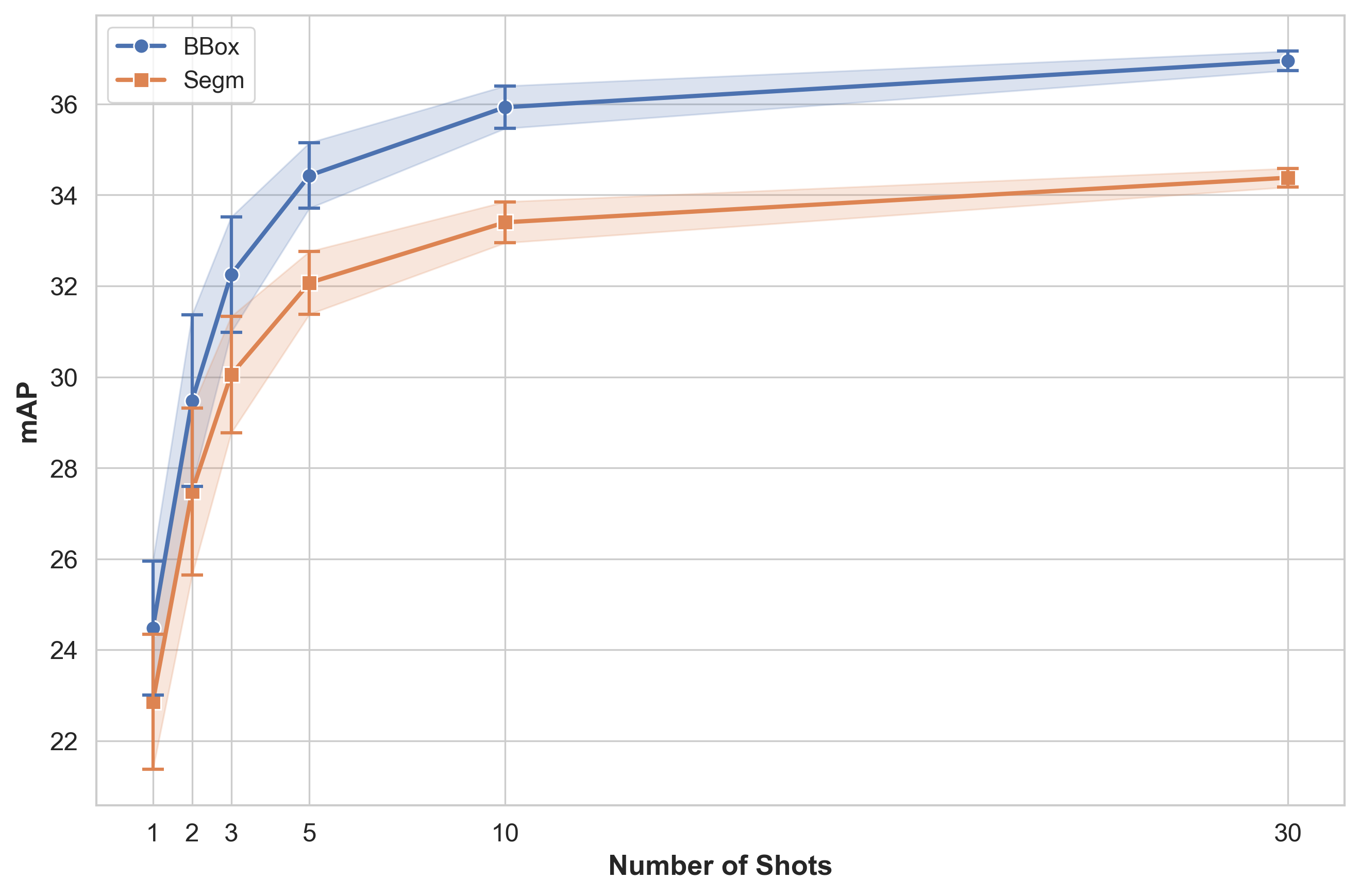}
  \caption{
  Variance (mAP) n-shot on COCO-20$^i$. Error bars show std over 10 runs with different reference image sets. Higher variance at low n-shot reflects sensitivity to reference images; it decreases as n increases, demonstrating our method’s robustness.
  }
  \label{fig:ablation-std}
\end{figure}
}

\subsubsection{Reference Masks Analysis}

We study three annotation characteristics that intuitively affect prototype quality: (1) mask area (object size), (2) mask center location, and (3) distance to image edges.

\textbf{Mask area.} Figure \ref{fig:mask-area-distribution} shows the distribution of mask areas. The distribution is skewed toward small masks, confirming that many annotations contain limited visual detail.

{\begin{figure}[h]
    \centering
    \includegraphics[width=0.75\linewidth]{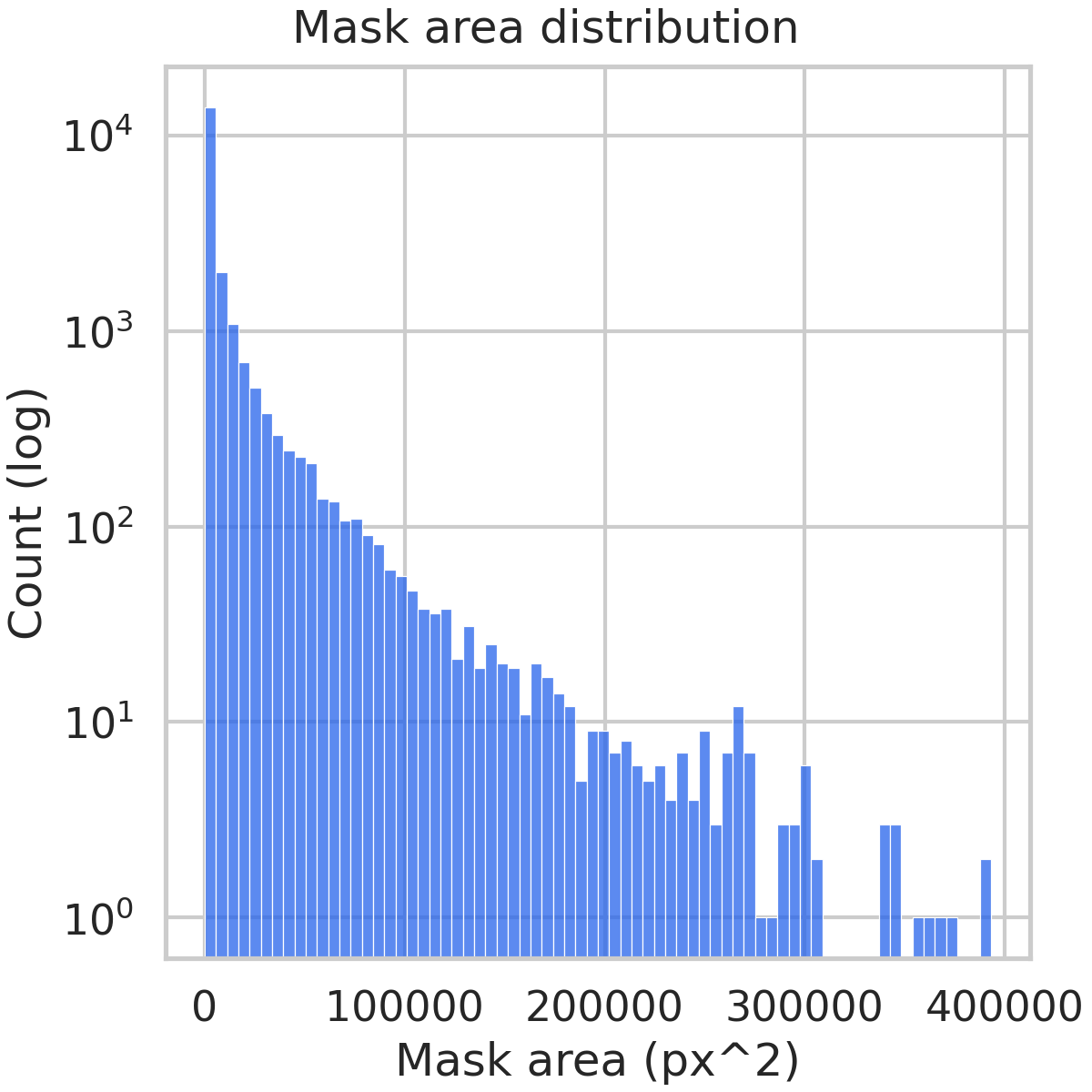}
    \caption{
    \textit{Mask-area distribution for the 20 COCO novel classes}. Most annotations are small or medium-sized, as shown by the positively skewed distribution (log-scale y-axis).}
    \label{fig:mask-area-distribution}
\end{figure}
}

\textbf{Mask center location.} Figure \ref{fig:grid-bbox-positions} shows 2D heatmaps of bounding-box centers. Most classes exhibit a centered bias, while some (e.g., car, chair) are more distributed along the horizontal axis.

\textbf{Mask distance to edge.} Figure \ref{fig:distance-to-frame} shows distance-to-edge histograms. A fraction of annotations overlap with image boundaries, implying that the object is partially cropped.

{\begin{figure}[h]
    \centering
    \includegraphics[width=\linewidth]{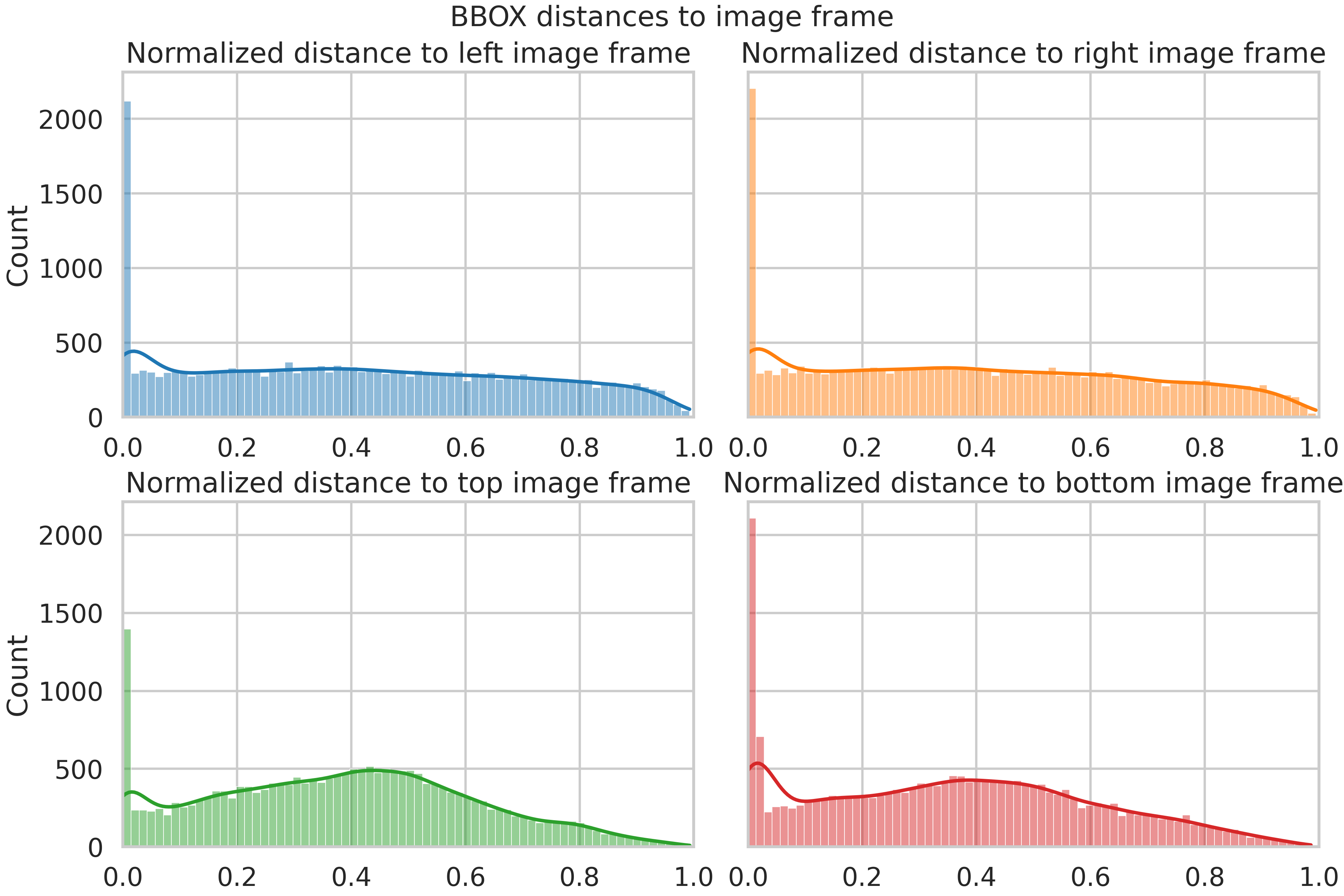}
    \caption{
    \textit{Distance from each mask to the four image boundaries}. A substantial number of annotations lie close to image edges, indicating possible partial cropping or truncated objects.}
    \label{fig:distance-to-frame}
\end{figure}
}

{\begin{figure*}[!htbp]
  \centering
  \includegraphics[width=0.95\linewidth]{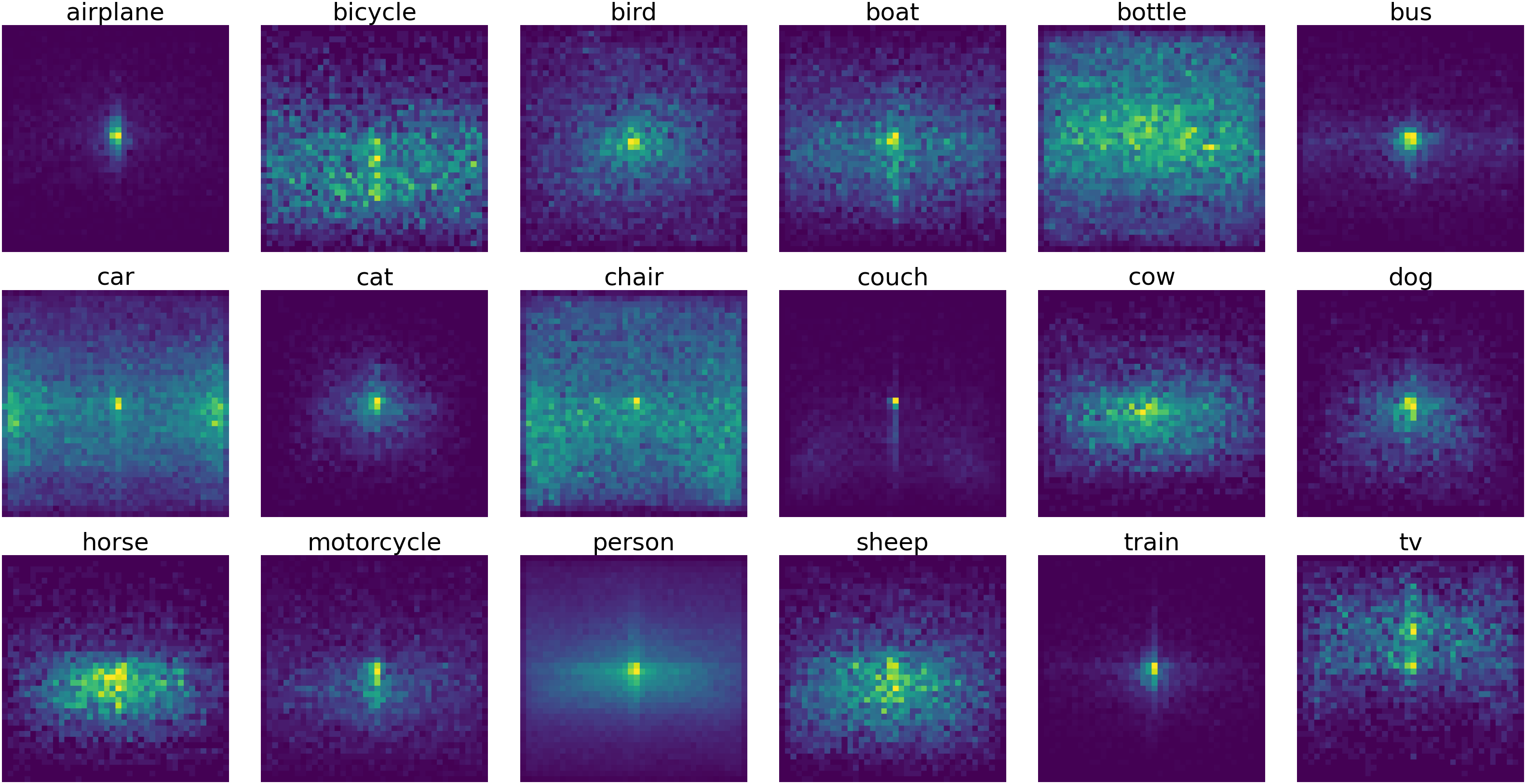}
  \color{blue}
  \caption{
  \textit{2D heatmaps of mask centers across COCO novel classes}. Most annotations are near the image center, though some classes show strong horizontal spread (e.g., car, chair).}
  \label{fig:grid-bbox-positions}
\end{figure*}
}
{    \begin{figure*}[!htbp]
      \centering
      \includegraphics[width=\linewidth]{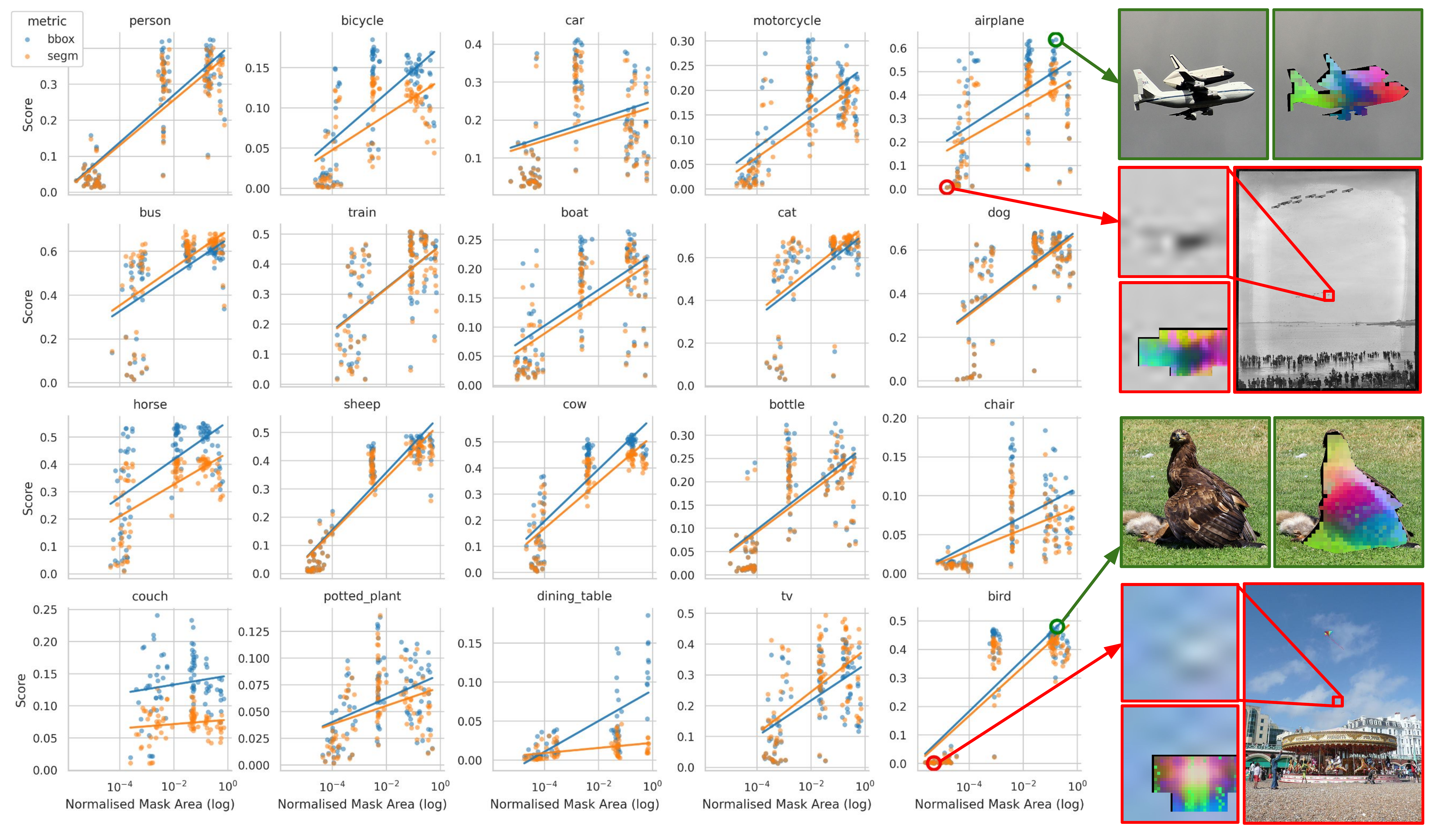}
      \color{blue}
      \caption{
      \textit{Reference-image performance vs. mask area for all COCO novel classes}. Each point represents one reference image evaluated on a class-specific reduced validation subset. Larger mask areas consistently correlate with higher scores. Right: visual examples comparing a high-performing and a low-performing reference for two classes. Best viewed when zoomed in.}
      \label{fig:class-performance}
    \end{figure*}
}

\subsubsection{Heuristics for Reference Selection}

In order to identify common patterns linked to higher performance, we sample 100 diverse reference images per class, explicitly covering a range of mask sizes, centers, and edge distances. Each reference is evaluated on a fixed reduced validation subset.

Figure \ref{fig:class-performance} shows per-class scatter plots of mask area vs. performance. All classes show a clear positive trend: larger masks consistently produce stronger reference prototypes which result in higher downstream scores.

Based on these observations we define simple selection heuristics:

\begin{itemize}[noitemsep=2pt, topsep=0pt]
    \item Area category (based on class-specific quartiles): large ($\geq$75th), medium (25th–50th), small ($\leq$25th).
    \item Centeredness: mask center within $\pm$10\% of the image center
    \item Edge avoidance: mask must be at least $d$ pixels away from any image boundary.
\end{itemize}

{\begin{figure}[h]
    \centering
    \includegraphics[width=0.9\linewidth]{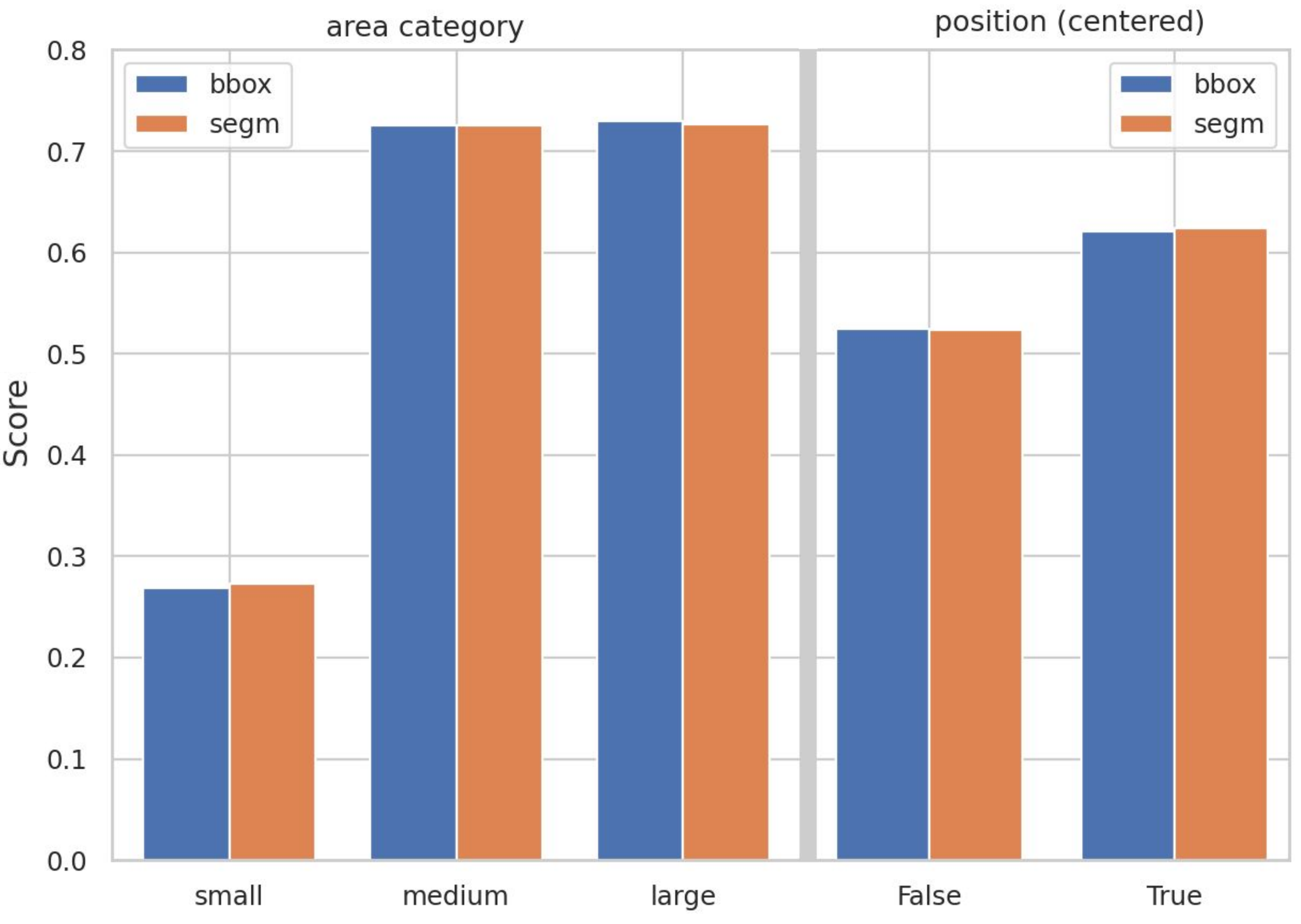}
    \caption{
    \textit{Effect of mask area (left) and centeredness (right) on downstream performance}. Medium/large masks significantly outperform small masks, and centered references outperform off-center ones.}
    \label{fig:barplot}
\end{figure}
}
{\begin{figure}[h]
    \centering
    \includegraphics[width=\linewidth]{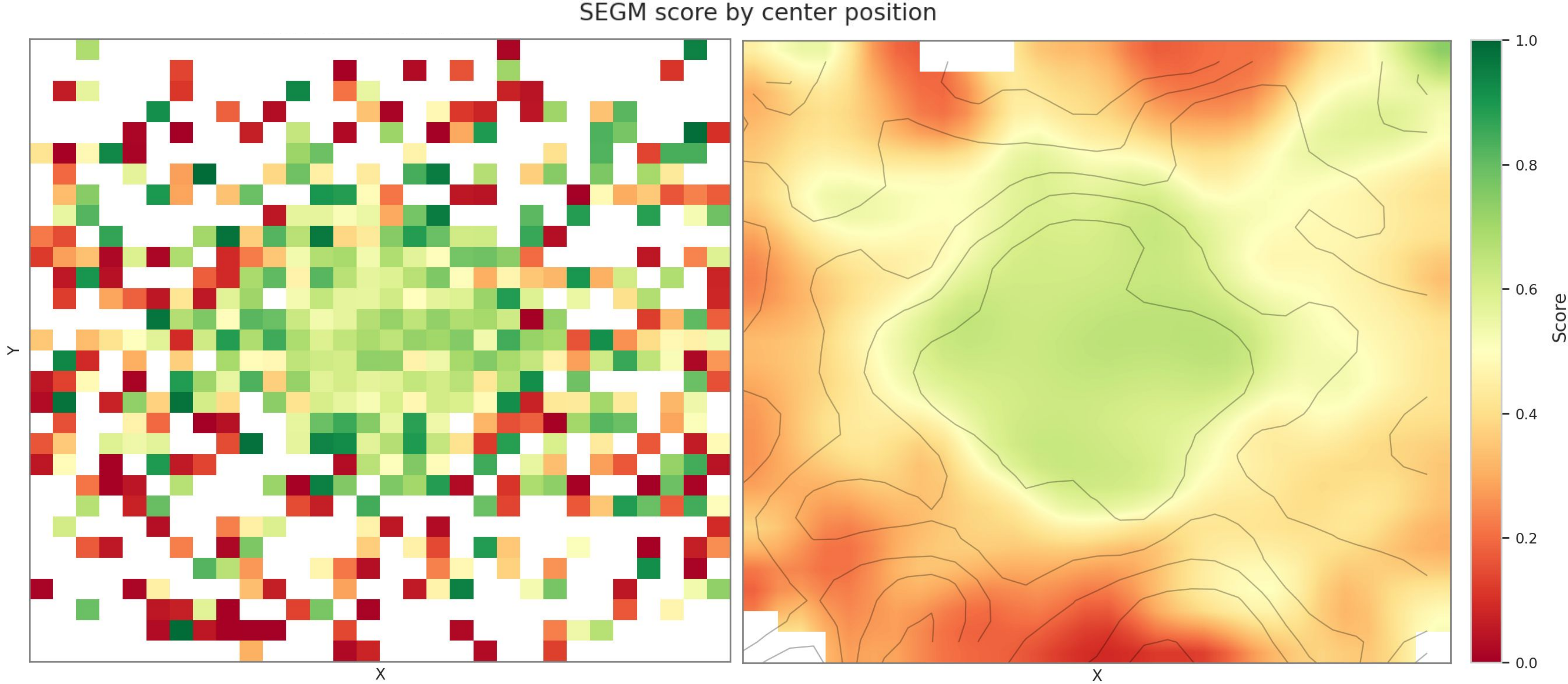}
    \caption{
    \textit{2D score maps of performance as a function of mask-center location}. Both the discrete and smoothed KDE versions show higher performance when reference masks lie near the image center.}
    \label{fig:heatmap}
\end{figure}
}

Figure \ref{fig:barplot} and Figure \ref{fig:heatmap} confirm that medium and large references outperform small ones, and centered references outperform off-center ones. Distance-to-edge has weaker effect once area and centeredness are controlled.

Overall, the proposed selection heuristics effectively improve one-shot performance without increasing the number of shots, providing a practical method for choosing high-value references.

\subsubsection{Reference-Image Degradation}
We evaluate our method under progressively degraded reference images by applying increasing levels of Gaussian blur. For efficiency, we conducted this experiment on a small representative subset, since our goal is to measure relative performance degradation rather than absolute metrics.

As shown in Figure \ref{fig:blur-ablation}, our method remains robust even under strong blur. We attribute this stability to the invariance and consistency of DINOv2 features, which remain discriminative despite significant image degradation.

{\begin{figure}[!htbp]
  \centering
  \includegraphics[width=0.9\linewidth]{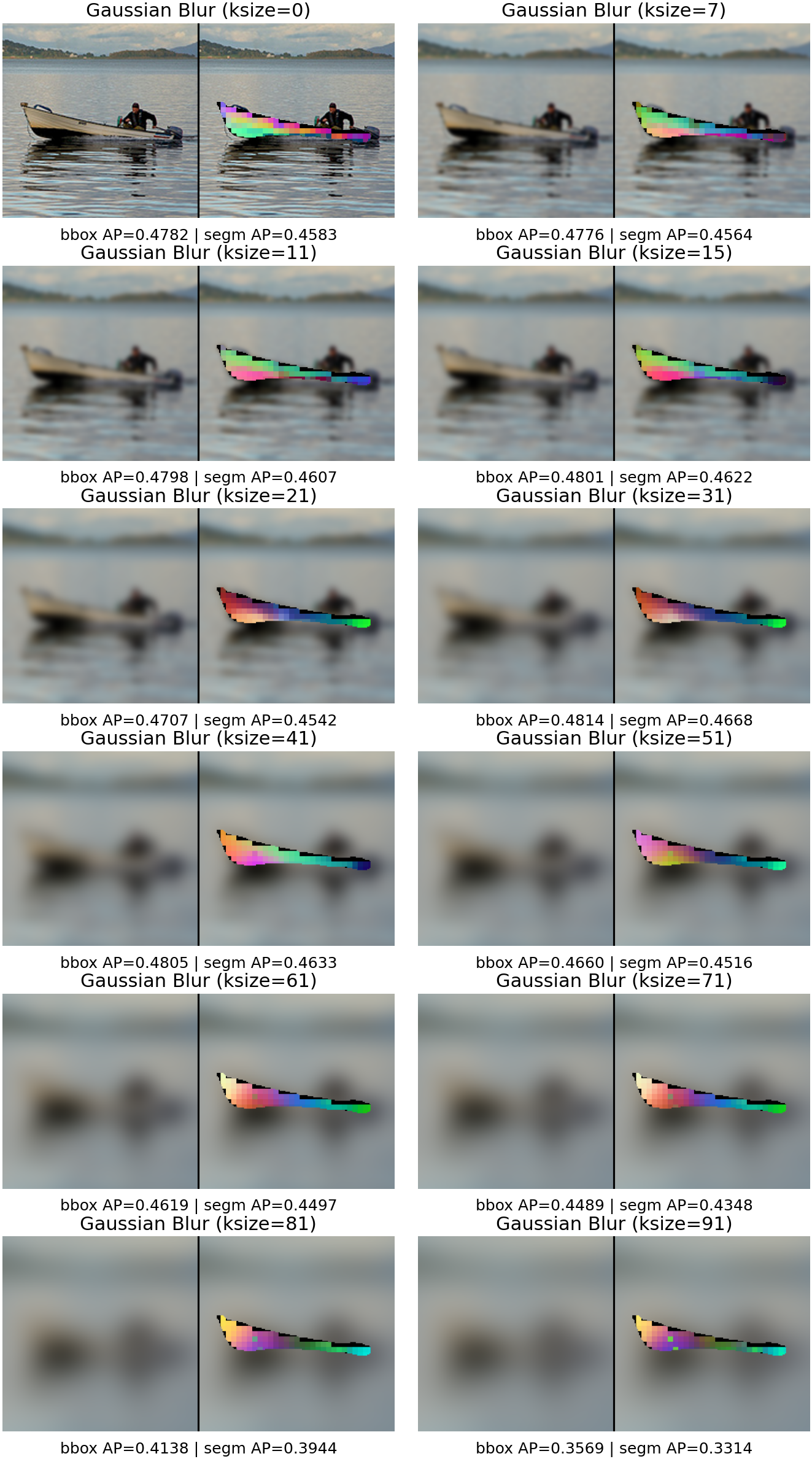}
  \color{blue}
  \caption{
  Model performance with increasing levels of blurring in the reference image. Each column shows a reference image blurred with an increasing Gaussian kernel size, along with its corresponding DINOv2 PCA feature visualization. Blurred input images result in blurred and smoother Dinov2 features. Below each pair, we display the predicted bounding box and segmentation mask. Our method exhibits strong robustness even under heavy blur.}
  \label{fig:blur-ablation}
\end{figure}
}

\subsection{Few-shot Semantic Segmentation on COCO}

Although our method is designed for instance segmentation, we also evaluate it on the COCO-20$^i$ Few-Shot Semantic Segmentation benchmark \cite{nguyen2019feature}. The 80 COCO classes are divided into four folds \cite{hu2019attention, panet, nguyen2019feature}, each containing 60 base and 20 novel classes. We assess performance on the 20 novel classes under strict 1-shot and 5-shot settings \cite{diffews}. Results are shown in Table \ref{tab:coco-semantic}. To adapt our instance segmentation predictions to semantic segmentation, we aggregate all instances of the same class into semantic maps, enabling direct comparison with prior methods. Despite being entirely \textit{training-free}, our method achieves competitive performance against fine-tuned approaches.

{

\begin{table}[h]
\centering
\begin{adjustbox}{width=\linewidth}
\addtolength{\tabcolsep}{-0.45em}
\begin{tabular}{lccccccccccc}
\toprule
\multicolumn{1}{c}{\multirow{2}{*}{\textbf{Methods}}} & \textbf{FT} & \multicolumn{5}{c}{\textbf{1-shot}} & \multicolumn{5}{c}{\textbf{5-shot}} \\
\cmidrule(lr){3-7} \cmidrule(lr){8-12}
& & $\mathbf{20^0}$ & $\mathbf{20^1}$ & $\mathbf{20^2}$ & $\mathbf{20^3}$ & \textbf{mean} & $\mathbf{20^0}$ & $\mathbf{20^1}$ & $\mathbf{20^2}$ & $\mathbf{20^3}$ & \textbf{mean} \\
\midrule
DiffewS~\cite{diffews}   & \cmark & 47.7 & 56.4 & 51.9 & 48.7 & 51.2 & 52.0 & 63.0 & 54.5 & 54.3 & 56.0 \\
DiffewS-n~\cite{diffews} & \cmark & 47.1 & 56.6 & 53.8 & 48.3 & 52.2 & 57.3 & 66.5 & 60.3 & 58.8 & 60.7 \\
\midrule
HSNet~\cite{hsnet}       & \xmark & 37.2 & 44.1 & 42.4 & 41.3 & 41.2 & 45.9 & 53.0 & 51.8 & 47.1 & 49.5 \\
CyCTR~\cite{cyctr}       & \xmark & 38.9 & 43.0 & 39.6 & 39.8 & 40.3 & 41.1 & 48.9 & 45.2 & 47.0 & 45.6 \\
VAT~\cite{vat}           & \xmark & 39.0 & 43.8 & 42.6 & 39.7 & 41.3 & 44.1 & 51.1 & 50.2 & 46.1 & 47.9 \\
BAM~\cite{bam}           & \xmark & 43.4 & 50.6 & 47.5 & 43.4 & 46.2 & 49.3 & 54.2 & 51.6 & 49.6 & 51.2 \\
DCAMA~\cite{dcama}       & \xmark & 49.5 & 52.7 & 52.8 & 48.7 & 50.9 & 55.4 & 60.3 & 59.9 & 57.5 & 58.3 \\
HDMNet~\cite{hdmnet}     & \xmark & 43.8 & 55.3 & 51.6 & 49.4 & 50.0 & 50.6 & 61.6 & 55.7 & 56.0 & 56.0 \\
\textbf{Training-free}   & \xmark & 32.1 & 46.6 & 50.9 & 48.7 & 44.6 & 55.2 & 49.5 & 60.7 & 45.5 & 52.7 \\
\bottomrule
\end{tabular}
\end{adjustbox}
\caption{Performance comparison of strict few-shot semantic segmentation settings (1 and 5-shot) on COCO-$20^i$. We aggregate instance-level predictions to allow comparison with semantic segmentation works. Previous methods results are sourced from \cite{diffews}. FT refers to finetuning on novel classes.}
\label{tab:coco-semantic}
\vspace{-5mm}
\end{table}
}

\subsection{Evaluating a VLM on COCO Few Shot Dataset}
\label{app:vlm}

To compare with recent VLM models, we implement a joint pipeline that combines Qwen2.5-VL-7B \cite{qwen} and SAM2. We run Qwen2.5-VL-7B on COCO validation set to produce category-conditioned bounding boxes, which are then directly used to prompt SAM2 to obtain segmentation masks. We evaluate both detection (nAP) and segmentation performance using the same COCO few-shot class split as our method. Table \ref{tab:ablation-vl-qwen} shows the results obtained, highlighting the performance of our model.

{\begin{table}[h]
\centering
\begin{adjustbox}{width=\linewidth}
\addtolength{\tabcolsep}{-0.4em}
\begin{tabular}{lccccccc}
\toprule
\multicolumn{1}{c}{\multirow{2}{*}{\makecell{\textbf{Backbone} \\ \textbf{models}}}} & \multicolumn{1}{c}{\multirow{2}{*}{\makecell{\textbf{Num} \\ \textbf{shots}}}} &  \multicolumn{3}{c}{\textbf{bbox}} & \multicolumn{3}{c}{\textbf{segm}} \\
\cmidrule(lr){3-5} \cmidrule(lr){6-8} 
& & \textbf{\small nAP} & \textbf{\small nAP50} & \textbf{\small nAP75} & \textbf{\small nAP} & \textbf{\small nAP50} & \textbf{\small nAP75} \\
\midrule
DINOv2-ViT-L-14 + SAM2-L & 30
    & 36.8 & 54.5 & 38.7 & 34.2 & 54.4 & 36.7 \\
Qwen2.5VL-7B-Inst + SAM2-L & -
    & 6.2 & 8.4 & 6.9 & 5.9 & 8.4 & 6.8 \\
\bottomrule
\end{tabular}
\end{adjustbox}
\caption{
Comparison between our DINOv2+SAM2 pipeline and the Qwen2.5-VL-7B+SAM2 baseline. Qwen-VL outputs bounding boxes that are fed to SAM2 for segmentation, while our method uses DINOv2 features with the same SAM2 decoder. Our approach achieves substantially higher detection and segmentation nAP across all metrics.}
\label{tab:ablation-vl-qwen}
\vspace{-5mm}
\end{table}
}

\subsection{Pipeline Design}
\label{app:pipeline-design}

We analyze two independent design choices in our pipeline:
(i) \textit{prototype construction}, which defines how reference features are aggregated into category prototypes; and
(ii) \textit{instance consolidation}, which determines how overlapping target predictions are combined into final instances.

\subsubsection{Prototype Construction}

We compare (1) class-only aggregation (direct feature averaging across all class pixels) and (2) our two-step strategy, where we first compute instance-level prototypes (pixel-weighted means within each mask) and then average them to obtain the class prototype.

As shown in Table \ref{tab:ablation-two-step-agg}, the performance difference between the two strategies is negligible. This is expected: when instances have similar pixel counts, both methods produce nearly identical prototypes. The two-step version is slightly more faithful to instance structure and therefore remains our default choice.

{\begin{table}[h]
\centering
\begin{adjustbox}{width=0.8\linewidth}
\addtolength{\tabcolsep}{-0.4em}
\begin{tabular}{lccccc}
\toprule
\multicolumn{1}{c}{\multirow{3}{*}{\makecell{\textbf{Backbone} \\ \textbf{models}}}} & \multicolumn{1}{c}{\multirow{3}{*}{\makecell{\textbf{Aggregation} \\ \textbf{strategy}}}} & \multicolumn{2}{c}{\textbf{10-shot}} & \multicolumn{2}{c}{\textbf{30-shot}} \\
\cmidrule(lr){3-4} \cmidrule(lr){5-6}
& & \multicolumn{1}{c}{\textbf{bbox}} & \multicolumn{1}{c}{\textbf{segm}} & \multicolumn{1}{c}{\textbf{bbox}} & \multicolumn{1}{c}{\textbf{segm}} \\
\cmidrule(lr){3-3} \cmidrule(lr){4-4} \cmidrule(lr){5-5} \cmidrule(lr){6-6}
& & \textbf{\small nAP} & \textbf{\small nAP} & \textbf{\small nAP} & \textbf{\small nAP} \\
\midrule
dinov2-sam2 & class agg
    & 35.7 & 33.4 
    & 36.8 & 34.4 \\
dinov2-sam2 & class+inst agg
    & 35.7 & 33.3
    & 36.8 & 34.2 \\
\bottomrule
\end{tabular}
\end{adjustbox}
\caption{
Ablation of two-step prototype aggregation strategy. We compare class-only aggregation (single-step) with two-step aggregation (instance-level pixel-weighted prototypes followed by class averaging). Results are reported for 10-shot and 30-shot settings using nAP, nAP50, and nAP75.}
\label{tab:ablation-two-step-agg}
\vspace{-5mm}
\end{table}
}

\subsubsection{Instance Consolidation}

We show the incremental improvements achieved by our \textbf{semantic-aware soft merging} in Table~\ref{tab:agg-strategies}.
Other aggregation variants, such as covariance similarity, instance softmax, score decay, iterative mask refine, attention-guided global average, underperformed.
Earth Mover’s Distance (EMD) was investigated in early design, but we found that its optimal-transport formulation introduced substantial computational overhead without providing clear qualitative benefits over cosine similarity.

\begin{table}[h]
\centering
\begin{adjustbox}{width=0.9\linewidth}
\addtolength{\tabcolsep}{-0.2em}
\setlength\extrarowheight{-2.5pt}
\begin{tabular}{lc}
\toprule
\textbf{Aggregation strategy} & \textbf{10-shot nAP} \\
\midrule
Hard-merging (hard threshold of 1 IoS)      & 31.2 \\
Soft-merging (without semantics)            & 35.7 \\
\textbf{Soft-merging (with semantics)}      & \textbf{36.6} \\
\bottomrule
\end{tabular}
\end{adjustbox}
\caption{Ablation on aggregation and matching strategies.}
\label{tab:agg-strategies}
\end{table}

\subsection{Feature Quality and Semantic Separability}
\label{app:feature-quality-separability}

\subsubsection{DINOv2 Feature Separability Analysis}

We compute DINOv2 features for all annotated objects in the COCO validation set (5k images) and visualize them using t-SNE. Figure \ref{fig:tsne} shows representative class pairs and triplets with varying degrees of semantic similarity.

Easily distinguishable classes (e.g., cat–dog) form well-separated clusters. In contrast, semantically similar categories (e.g., car–truck, chair–couch, wine glass–cup–vase) exhibit substantial embedding overlap. This indicates that similar-class confusion is driven by DINOv2’s feature geometry rather than prototype construction.

Improving performance in such cases requires stronger semantic disentanglement at the backbone level, which remains an open research direction.

{\begin{figure*}[h]
  \centering
  \includegraphics[width=\linewidth]{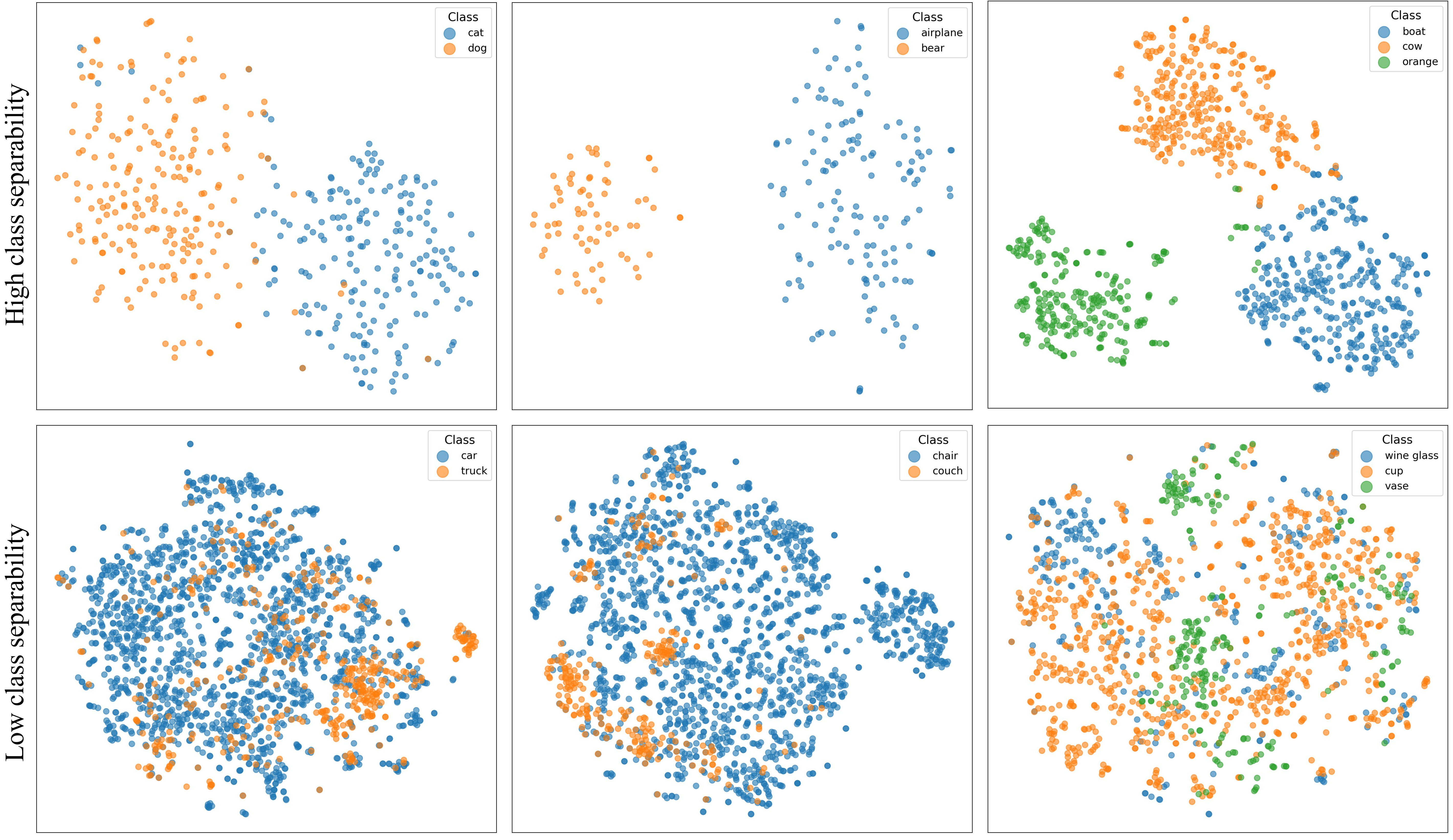}
  \color{blue}
  \caption{
  \textit{t-SNE visualizations of DINOv2 object-level features for selected COCO classes.} Top row: pairs of classes with clear separability (cat–dog, airplane–bear, boat–cow–orange). Bottom row: semantically similar classes with substantial embedding overlap (car–truck, chair–couch, wine glass–cup–vase). These results indicate that similar-class confusion primarily originates from DINOv2’s feature-space entanglement, limiting the effectiveness of alternative prototype-construction strategies, and setting an upper bound for our method performance.}
  \label{fig:tsne}
\end{figure*}
}

\subsubsection{Backbone comparison: DINOv2 vs. DINOv3}

\textbf{Feature separability.}
Figure \ref{fig:tsne-backbones} shows t-SNE embeddings for DINOv2-L and DINOv3-B/L/H. Classes that are not well-separated in DINOv2 remain similarly entangled in all DINOv3 variants, indicating that the semantic structure relevant for prototype-based matching does not substantially improve in DINOv3. Figure \ref{fig:feature-backbone-comparison} provides complementary PCA visualisations of the spatial feature maps, showing that DINOv3 produces smoother and more spatially uniform features, while DINOv2 exhibits sharper local variation. These results suggest that although DINOv3 representations are denser, their object-level semantics remain comparable to DINOv2, which may limit performance improvements in our training-free setting.

\textbf{Backbone size effects.}
Figure \ref{fig:output-dino-comparison} compares model outputs across backbones. DINOv3 backbones (B/L/H) tend to assign high similarity scores more broadly across proposals.

{\begin{figure*}[h]
  \centering
  \includegraphics[width=0.95\linewidth]{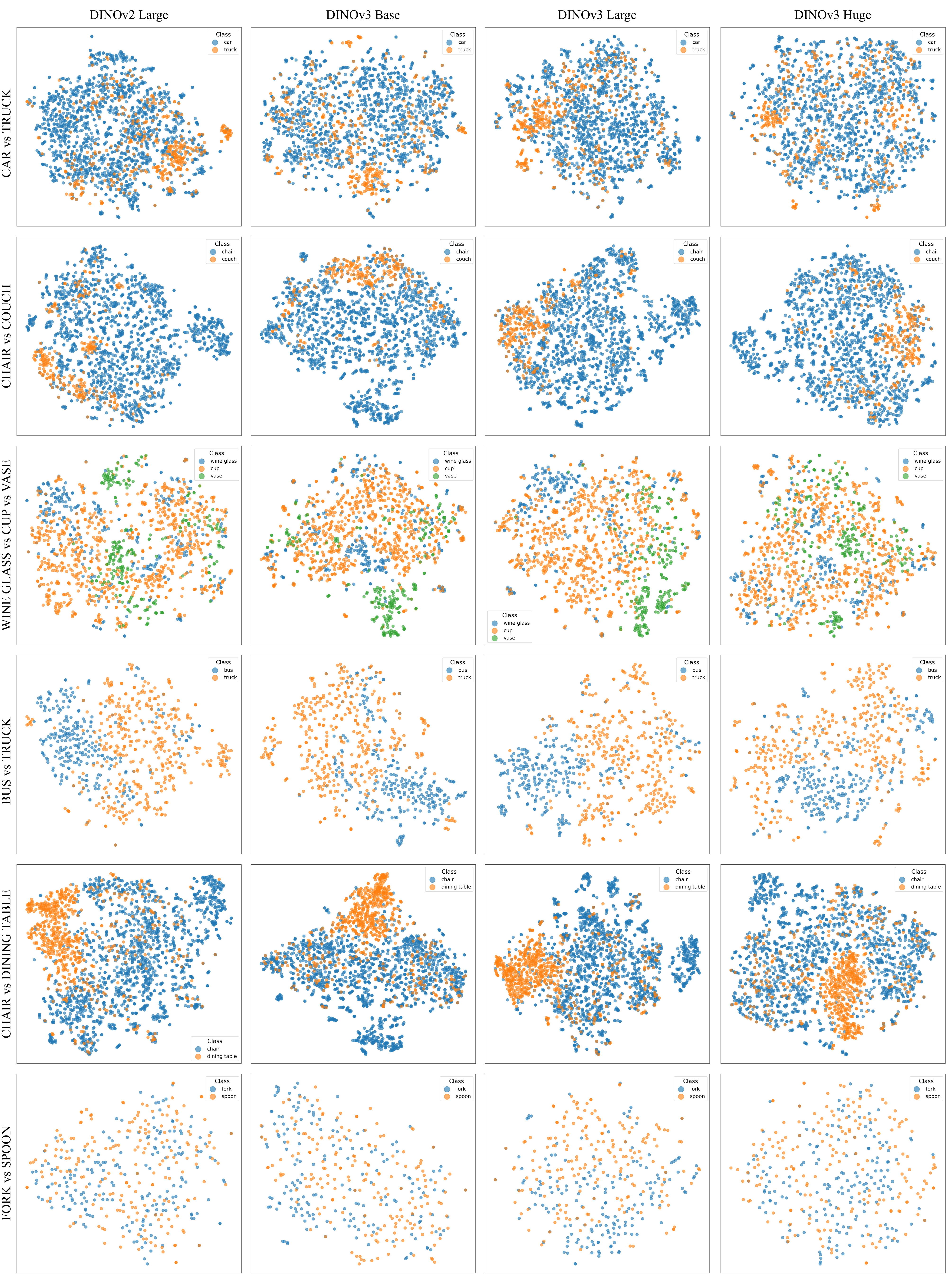}
  \color{blue}
  \caption{
  \textit{t-SNE embeddings across DINO backbones.} t-SNE visualisation of feature embeddings for representative classes (e.g., fork and spoon) using DINOv2-L, DINOv3-B, DINOv3-L, and DINOv3-H. Clusters that overlap in DINOv2 remain overlapping in DINOv3 variants, indicating that semantic separability does not improve across these models.}
  \label{fig:tsne-backbones}
\end{figure*}
}

{\begin{figure*}[h]
  \centering
  \includegraphics[width=0.9\linewidth]{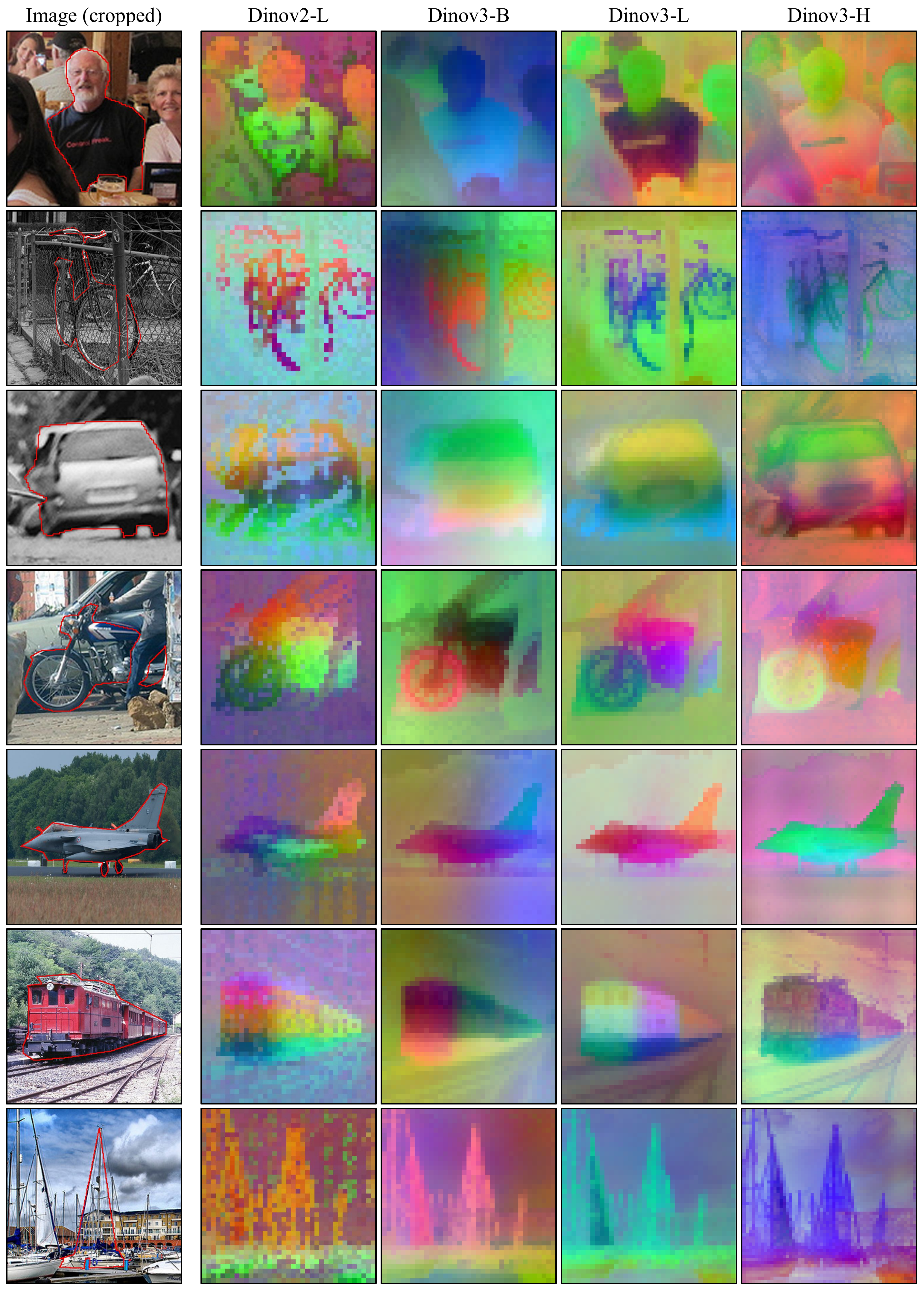}
  \color{blue}
  \caption{
  \textit{Feature visualisations across DINO variants.} PCA projections of the spatial features for DINOv2-L and DINOv3-B/L/H.}
  \label{fig:feature-backbone-comparison}
\end{figure*}
}

{\begin{figure*}[h]
  \centering
  \includegraphics[width=0.73\linewidth]{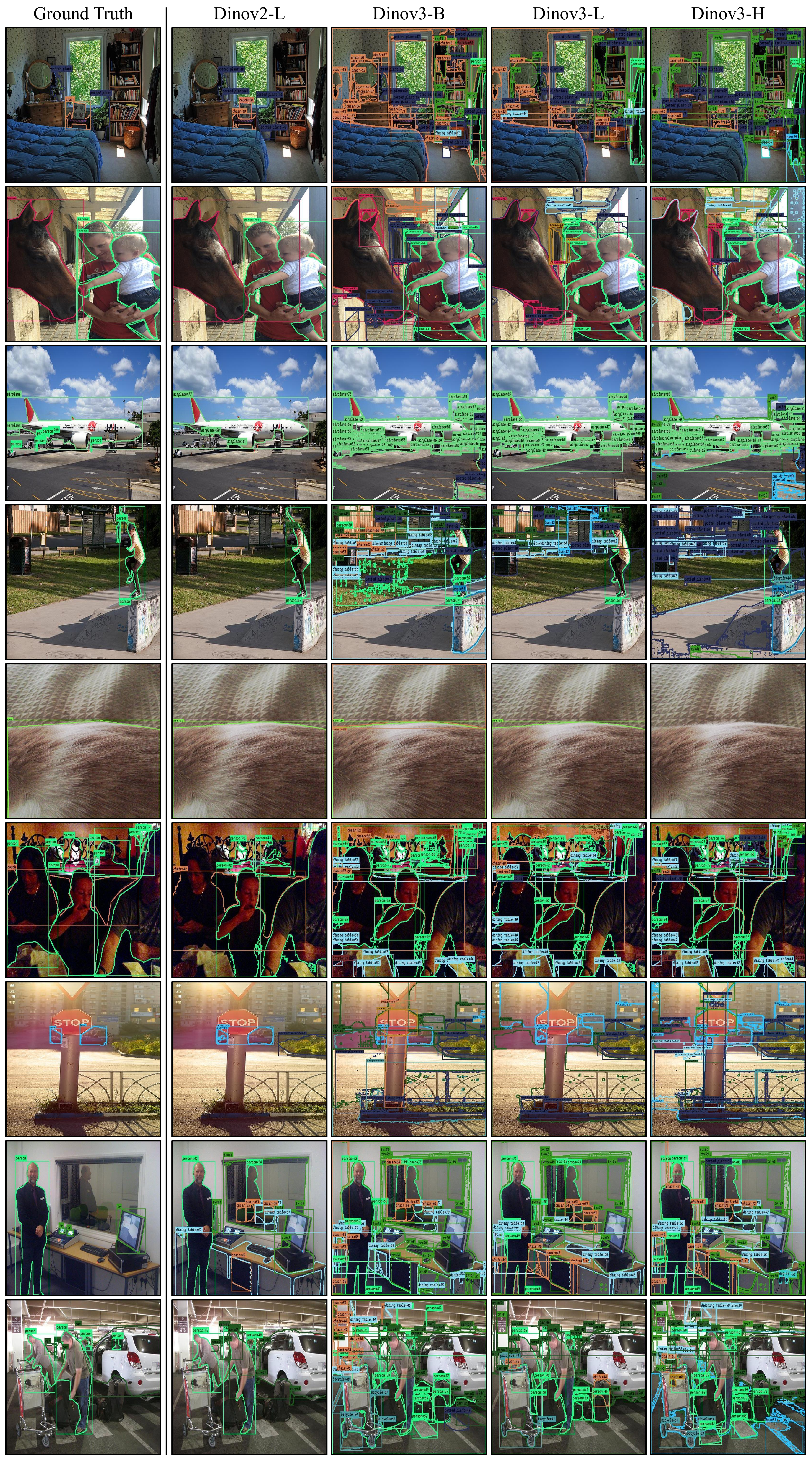}
  \color{blue}
  \caption{
  \textit{Output comparison across DINOv3 backbones.} Comparison of training-free outputs for DINOv2-L and DINOv3-B/L/H. In DINOv3 models, higher similarity scores are often assigned to multiple proposals, and predictions show less object-focused contrast. Ground-truth masks are shown for reference.}
  \label{fig:output-dino-comparison}
\end{figure*}
}

\subsection{Implementation details}
\label{app:implementation}

\textbf{Inference Runtime Efficiency.}
Our training-free stages are optimised and lightweight.
(1) Memory bank construction (0.1 s/img) is computed once, only requiring to encode $n$ reference images per class with DINOv2. Reference-image features are pre-cached for all following steps.
(2) Semantic matching (0.0003 s/img) is a fully parallelised dot product of cosine similarities.
(3) Soft-merging (0.006 s/img) uses a parallel implementation of NMS.
Table~\ref{tab:time-per-image} shows our method yields significant performance gains compared to Matcher \cite{Liu2023May_matcher} and speeds up SAM default automatic mask generator (AMG) by $\times3$ via efficient point sampling, faster mask filtering, and removal of unnecessary post-processing.
\\

{\begin{table}[h]
\centering
\begin{adjustbox}{width=0.9\linewidth}
\addtolength{\tabcolsep}{-0.2em}
\setlength\extrarowheight{-1pt}
\begin{tabular}{lc}
\toprule
\textbf{Method} & \textbf{Time (sec/img)} \\
\midrule
Matcher~[45]                                & 120.014 \\
Training-free (ours) with SAM AMG     & 3.5092 \\
\textbf{Training-free (ours)}               & \textbf{0.9292} \\
\bottomrule
\end{tabular}
\end{adjustbox}
\caption{Time to process an image on 20 ref. classes with A100.}
\label{tab:time-per-image}
\vspace{-5mm}
\end{table}
}

\textbf{Model Size and Memory Usage.}
We report the runtime footprint of our pipeline when using 20 classes and 10 shots per class (200 reference images). At inference time, the combined model size is 1.97 GiB, consisting of the DINOv2 encoder (1.13 GiB) and the SAM2 predictor (856 MiB). CUDA-allocated memory reaches a peak of 11.5 GiB. These measurements provide a reference for practical deployment and indicate potential avenues for integrating lighter variants (e.g., MobileSAM~\cite{mobilesam} or LightSAM~\cite{lightsam}) when targeting edge devices.

\subsection{Additional figures}

In Figure \ref{fig:feature-comparison-large}, we provide extended examples comparing the PCA projections of DINOv2, DINOv3, CLIP, and PE-Spatial features, showing spatial structure and noise characteristics across backbones.

Figure \ref{fig:output-comparison-large} includes extended model output visualizations for semantic backbones, highlighting how differences in feature quality propagate to detection and segmentation predictions.

Figures \ref{fig:feature-similarity-large} and \ref{fig:feature-similarity-large-extras} contain examples of similarity maps for both single-feature and aggregated-feature settings across more images and classes.

{\begin{figure*}[h]
  \centering
  \includegraphics[width=0.88\linewidth]{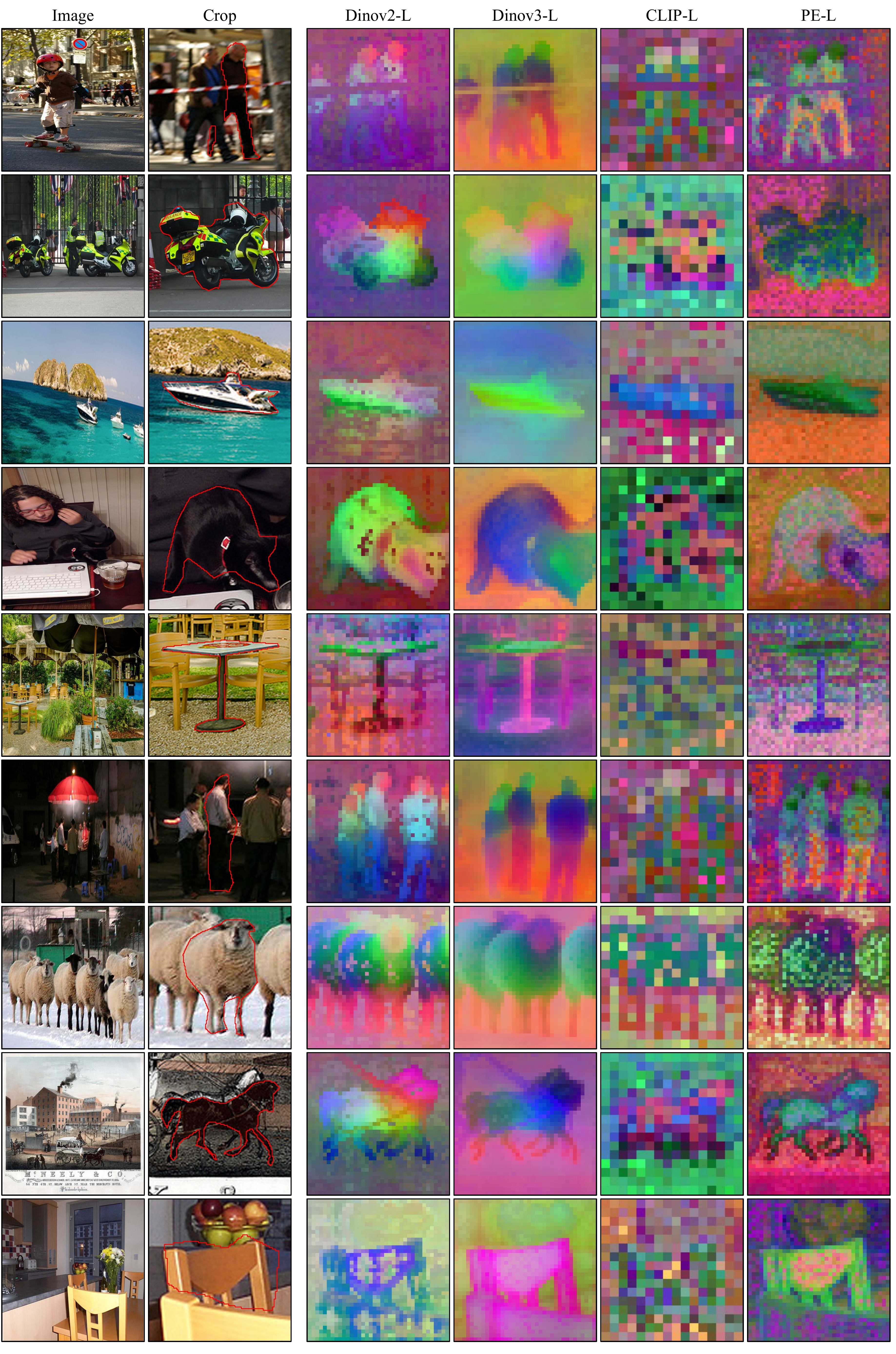}
  \color{blue}
  \caption{
  \textit{Extended feature comparison across semantic backbones.} PCA projections of DINOv2, DINOv3, CLIP, and PE-Spatial feature maps for additional examples. These visuals highlight qualitative differences in spatial resolution, noise, and feature consistency.}
  \label{fig:feature-comparison-large}
\end{figure*}
}
{\begin{figure*}[h]
  \centering
  \includegraphics[width=0.95\linewidth]{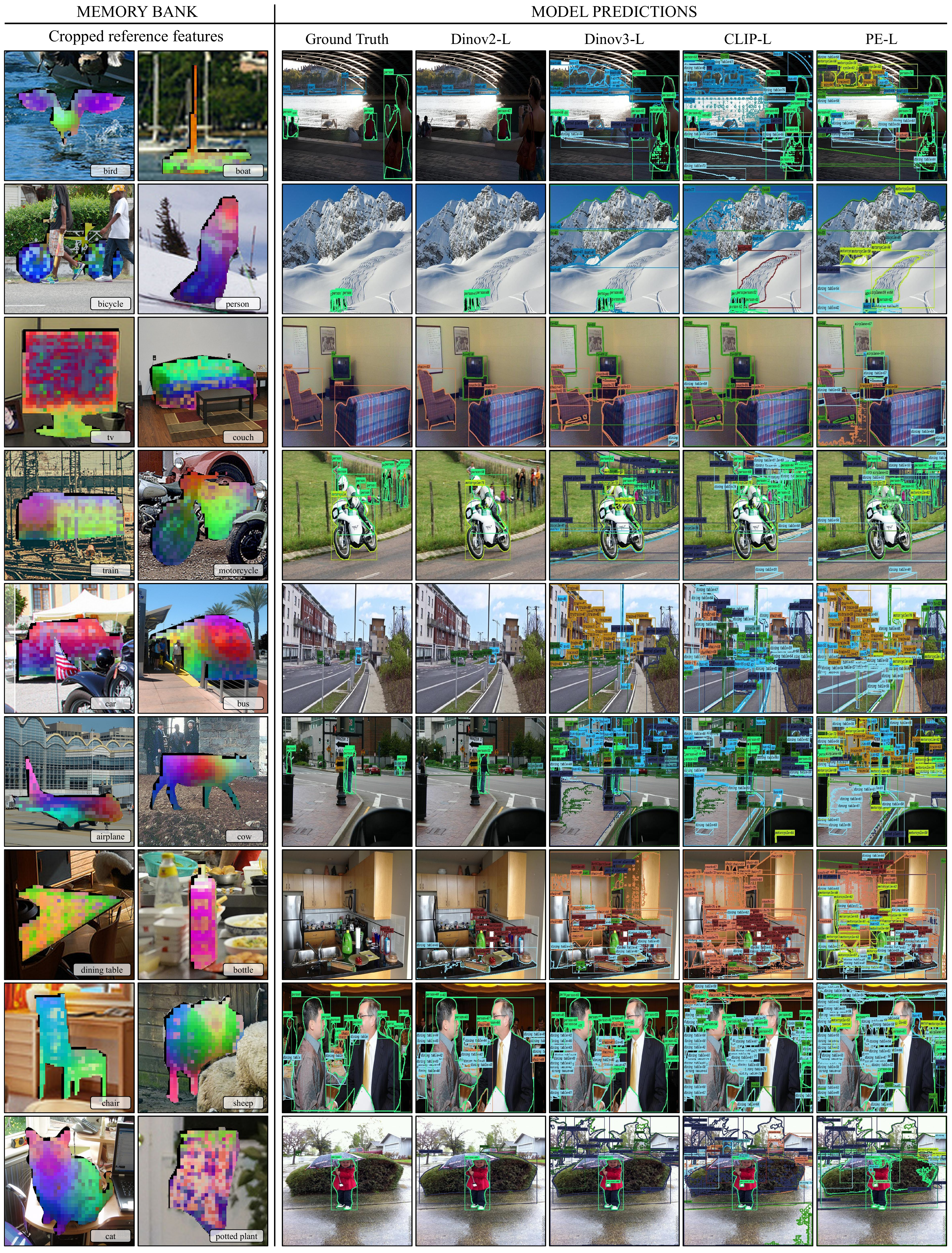}
  \color{blue}
  \caption{\textit{Extended output comparison across semantic backbones.} Predictions from multiple backbones on additional images, including ground-truth masks and corresponding model outputs. Differences in feature quality lead to distinct segmentation and detection behaviours. Best viewed when zoomed in.}
  \label{fig:output-comparison-large}
\end{figure*}
}
{\begin{figure*}[h]
  \centering
  \includegraphics[width=0.8\linewidth]{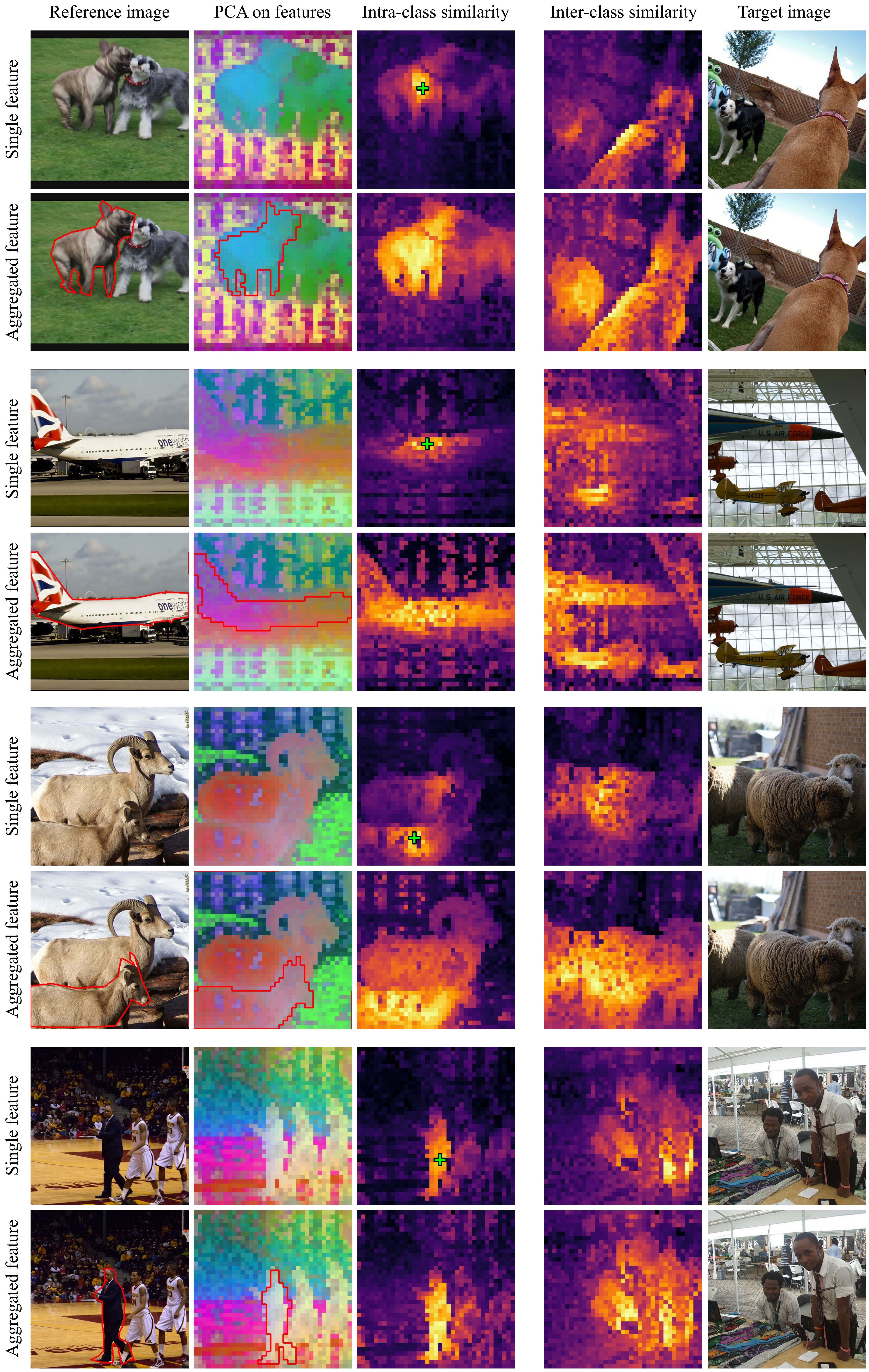}
  \color{blue}
  \caption{\textit{Feature similarity — additional paired examples.} For each reference-target image pair: the top row shows cosine similarity maps computed from a single DINOv2 patch feature (marked {\contour{black}{\textcolor{green}{\textbf{+}}}}), and the bottom row shows maps computed from the aggregated prototype (mask-area average). For each row we display intra-class (same image) and inter-class (target image) similarity. Paired examples demonstrate that aggregated prototypes yield more coherent, object-level similarity than single patch features.}
  \label{fig:feature-similarity-large}
\end{figure*}
}
{\begin{figure*}[h]
  \centering
  \includegraphics[width=0.8\linewidth]{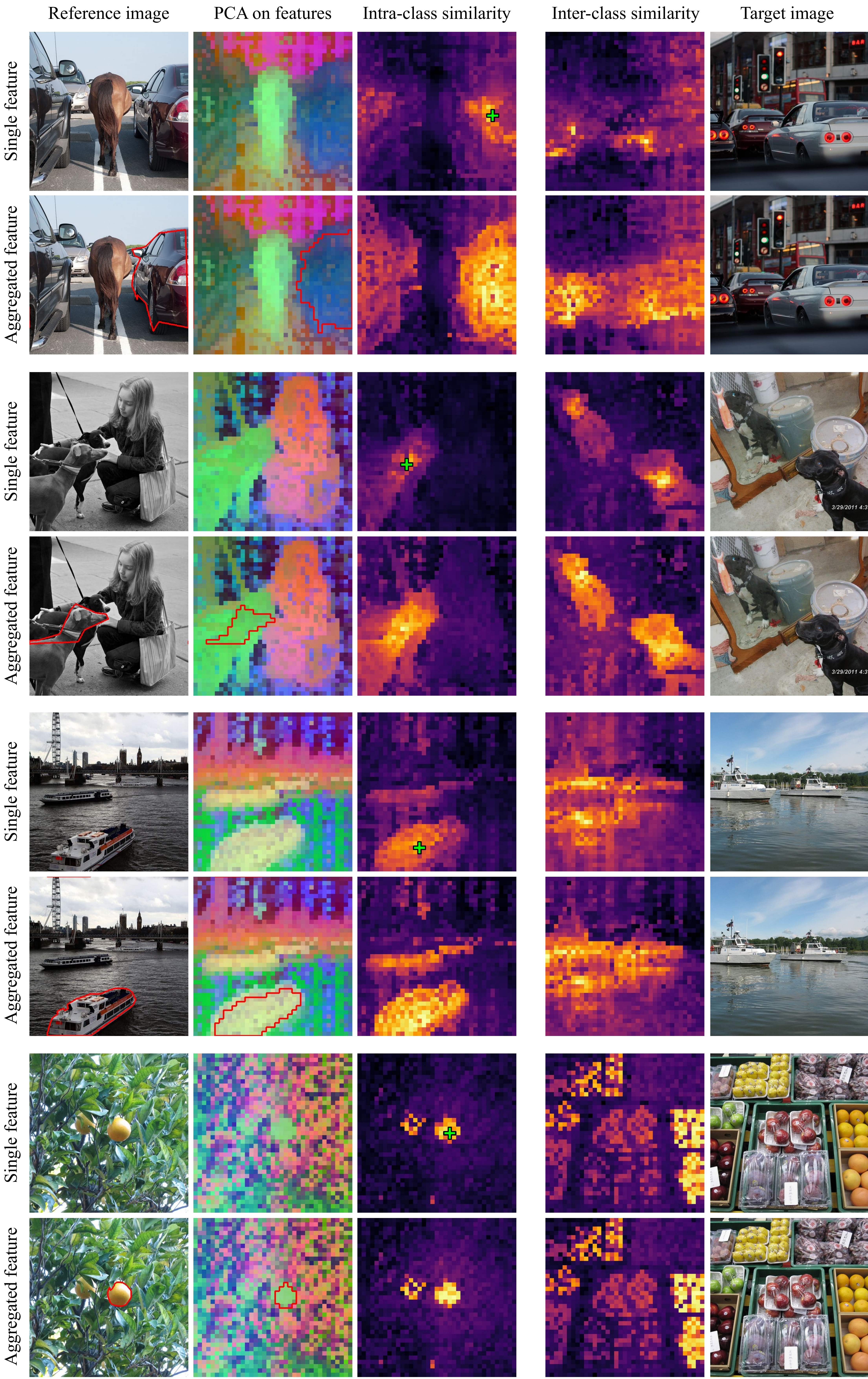}
  \color{blue}
  \caption{\textit{Feature similarity — extended paired examples.} For each reference-target image pair: the top row is single-feature similarity (patch {\contour{black}{\textcolor{green}{\textbf{+}}}}) and the second row is prototype similarity (mask-average). Each pair shows intra- and inter-class cosine similarity. These examples reinforce that aggregation produces more stable and spatially consistent instance-level matches.}
  \label{fig:feature-similarity-large-extras}
\end{figure*}
}

\subsection{Cross-Domain Few-Shot Object Detection}
\label{app:cdfsod}

{\begin{figure*}[h]
  \centering
  \includegraphics[width=0.9\linewidth]{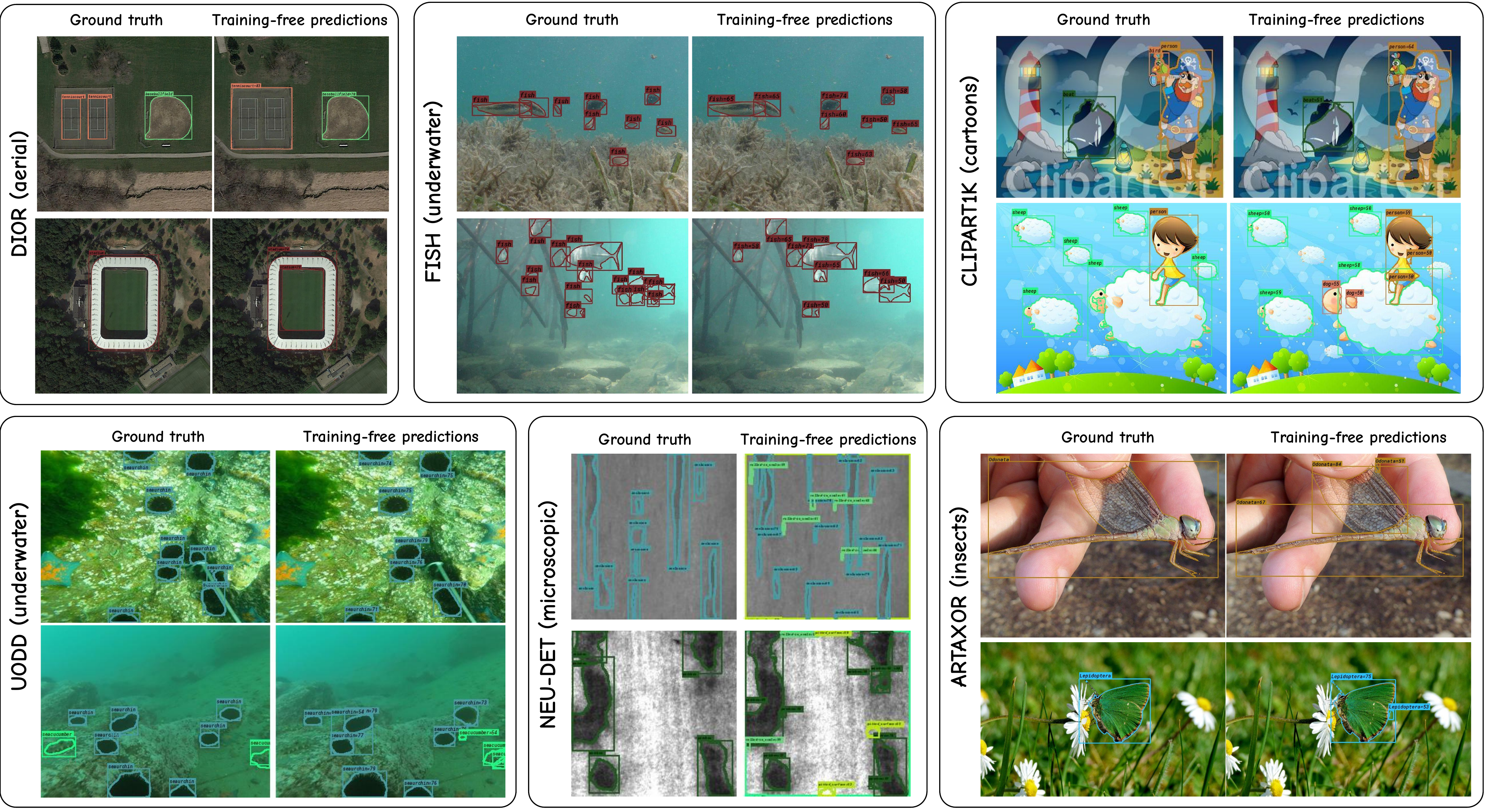}
  \caption{\textbf{Cross-domain \underline{5-shot} segmentation results using our training-free method.} Our approach evaluates diverse datasets across multiple domains, including aerial, underwater, microscopic, and cartoon imagery, without requiring fine-tuning. Results demonstrate the robustness and generalisability of our method.
 }
  \label{fig:cdsod-5-shot}
\end{figure*}
}

We provide additional visualisations of the six target datasets in the CD-FSOD benchmark. These datasets span diverse and challenging domains, including photorealistic, cartoon, aerial, underwater, and industrial imagery, each presenting unique distribution shifts. Despite these variations, our method achieves strong performance across all domains without any fine-tuning, demonstrating its remarkable cross-domain generalization. Figure \ref{fig:cdsod-5-shot} shows an overview of the results for the 6 datasets. Figures \ref{fig:artaxor}, \ref{fig:uodd}, \ref{fig:clipart1k}, \ref{fig:fish}, \ref{fig:dior}, \ref{fig:neu-det} display more detailed results for each of the datsets. All results are shown for 5-shot setting, using 5 reference images per category.

{\begin{figure*}
  \centering
  \includegraphics[width=\linewidth]{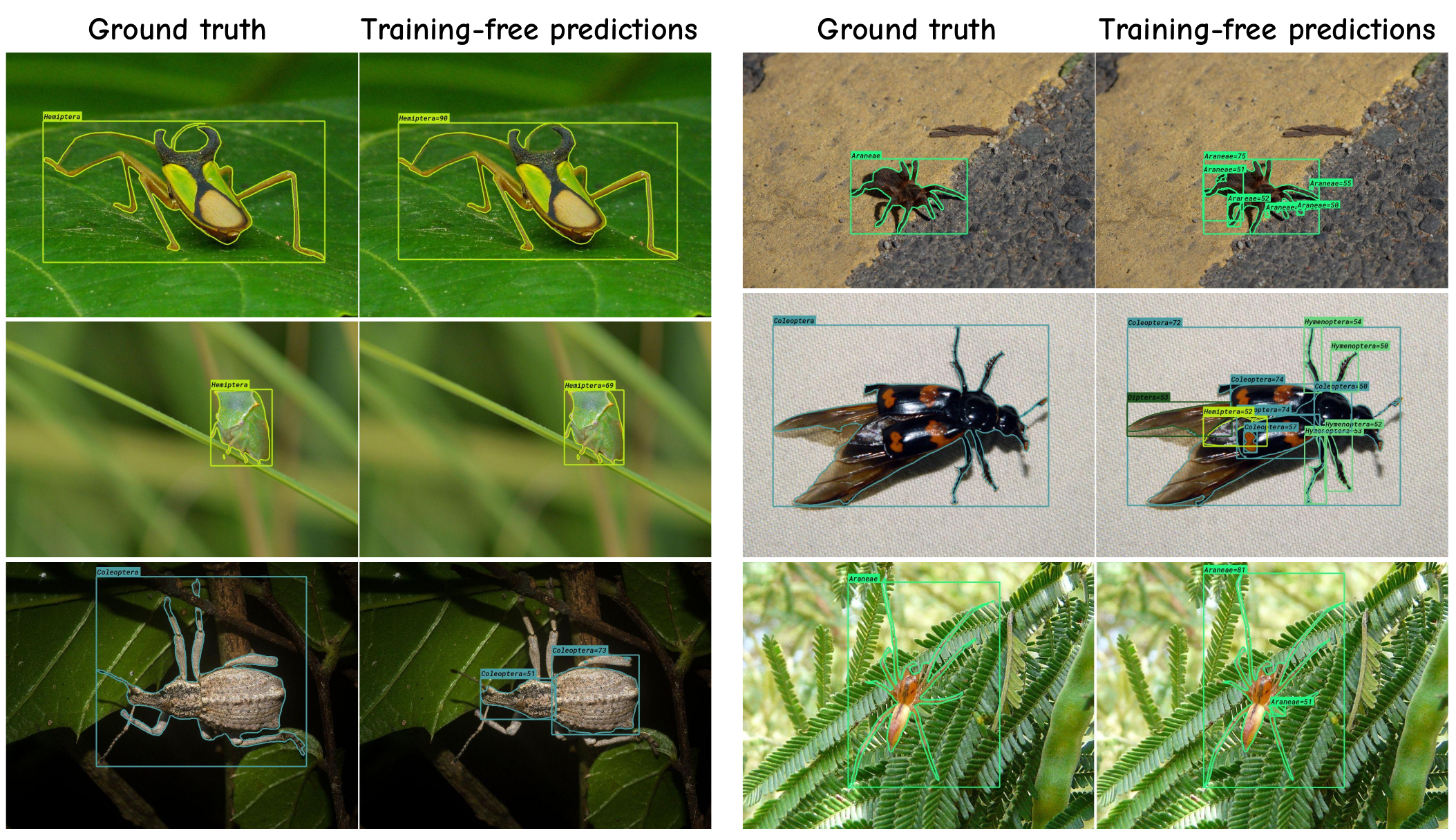}
  \caption{5-shot results on the ArTaxOr dataset.}
  \label{fig:artaxor}
\end{figure*}
}
{\begin{figure*}
  \centering
  \includegraphics[width=0.93\linewidth]{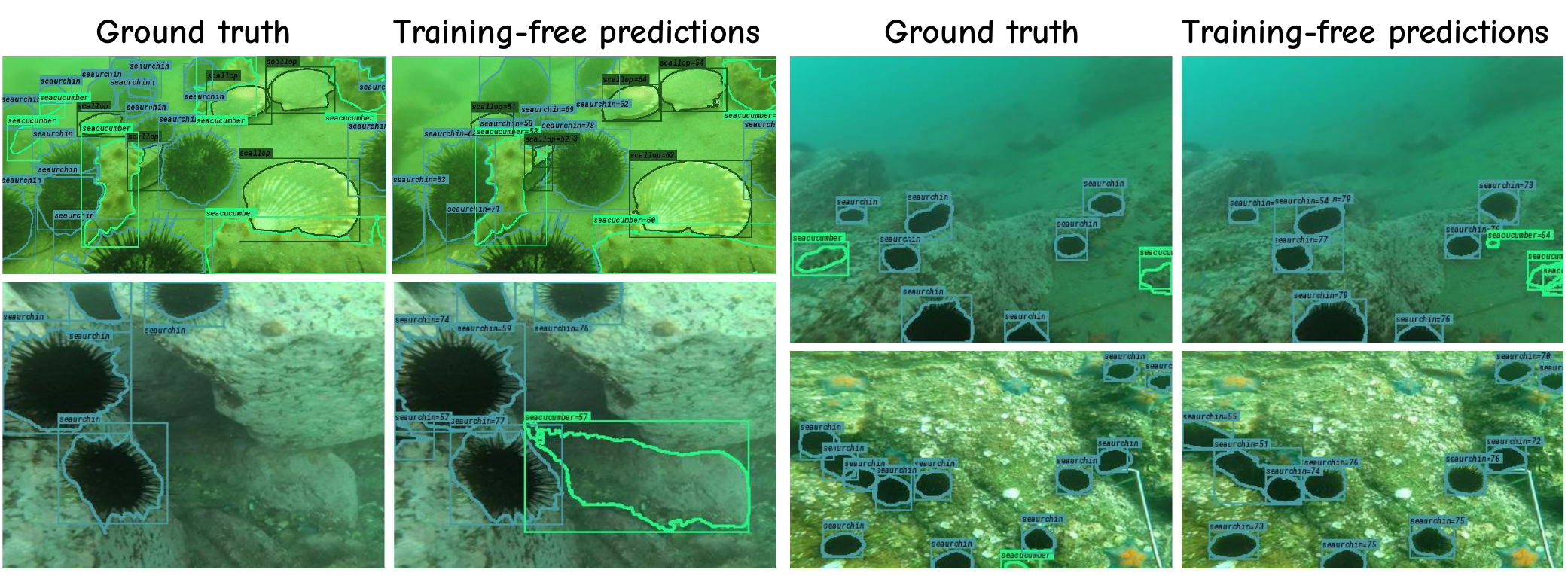}
  \caption{5-shot results on the UODD underwater dataset.}
  \label{fig:uodd}
  \vspace{-0.4cm}
\end{figure*}
}
{\begin{figure*}
  \centering
  \includegraphics[width=0.93\linewidth]{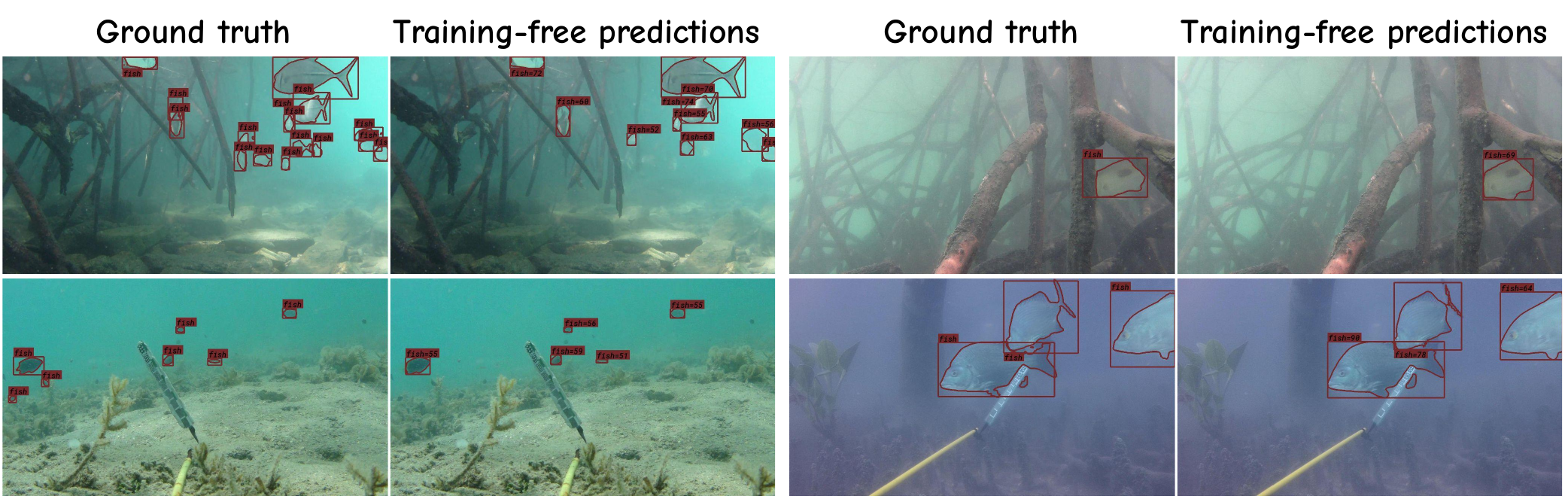}
  \caption{5-shot results on the Fish dataset.}
  \label{fig:fish}
  \vspace{-0.4cm}
\end{figure*}
}
{\begin{figure*}
  \centering
  \includegraphics[width=0.93\linewidth]{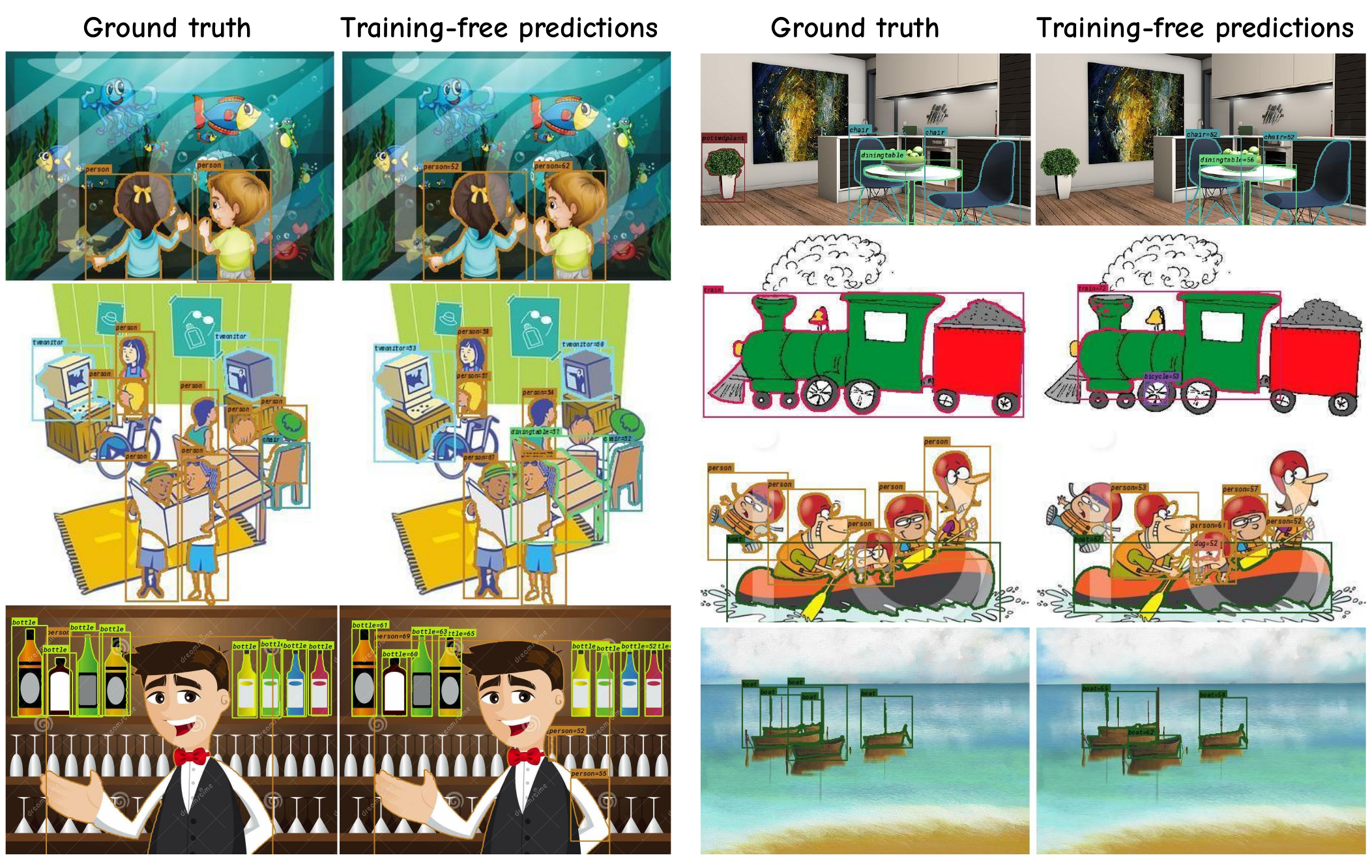}
  \caption{5-shot results on the Clipart1k dataset.}
  \label{fig:clipart1k}
  \vspace{-0.4cm}
\end{figure*}
}
{\begin{figure*}
  \centering
  \includegraphics[width=0.93\linewidth]{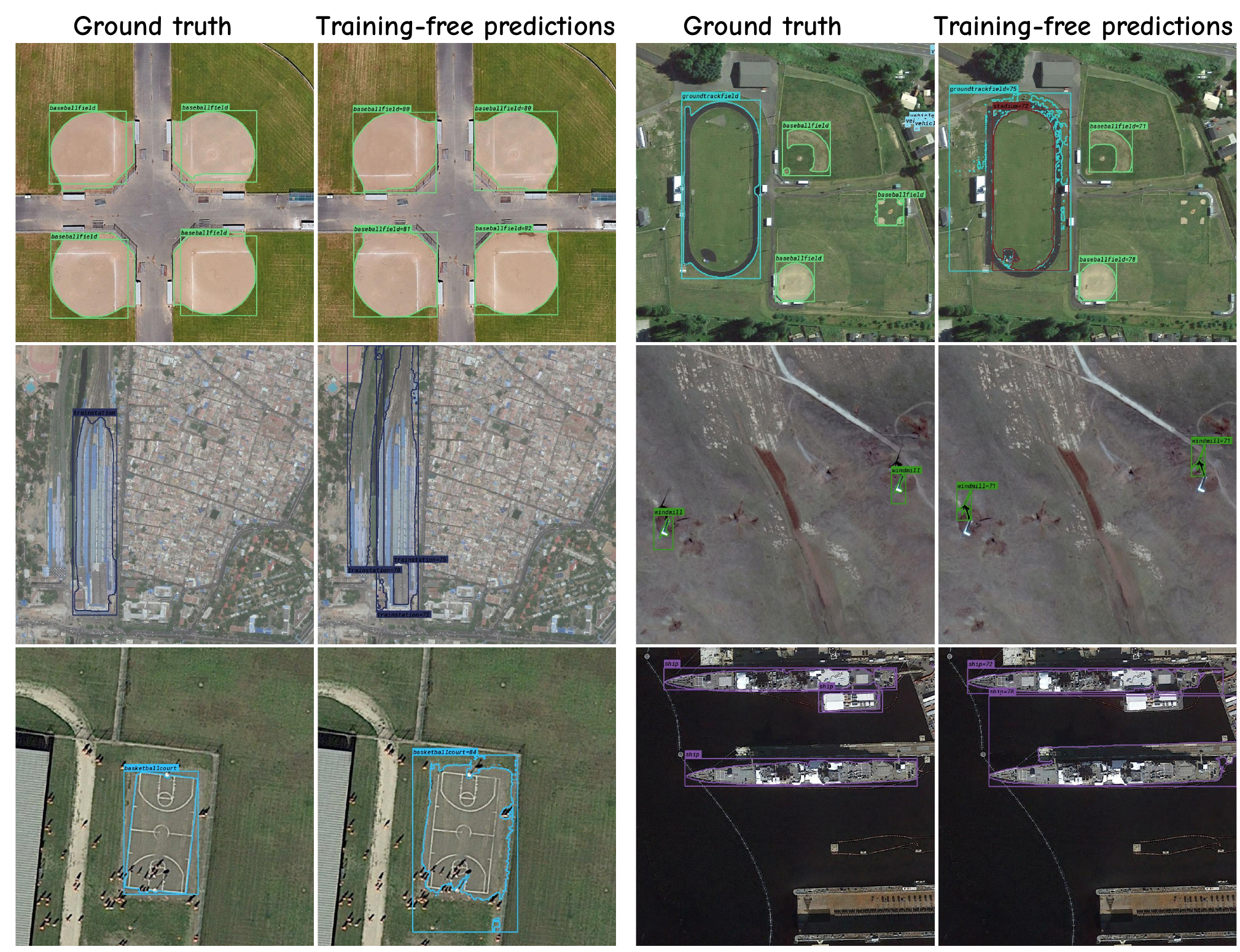}
  \caption{5-shot results on the DIOR dataset.}
  \label{fig:dior}
  \vspace{-0.4cm}
\end{figure*}
}
{\begin{figure*}
  \centering
  \includegraphics[width=0.93\linewidth]{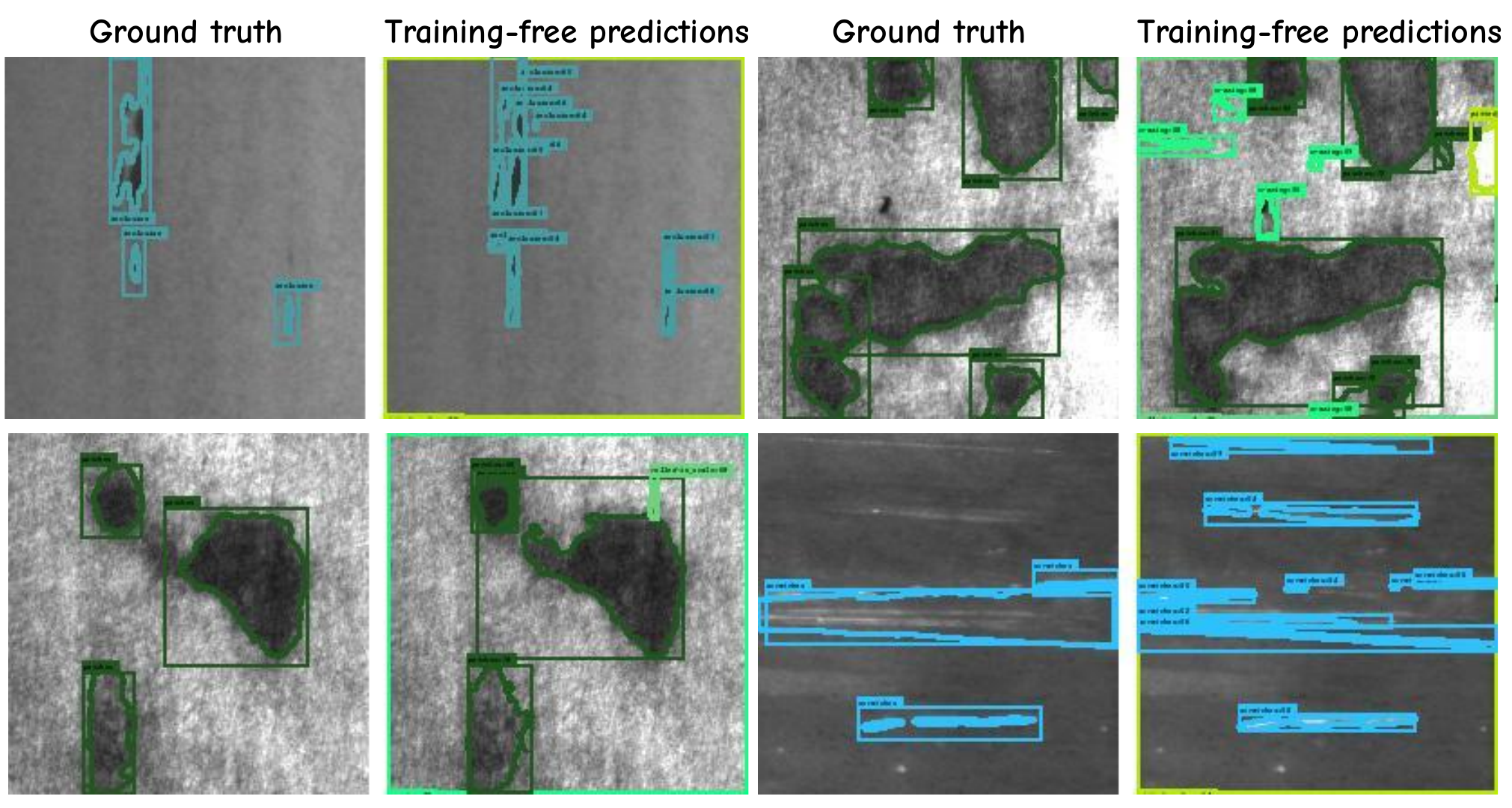}
  \caption{5-shot results on the NEU-DET dataset.}
  \label{fig:neu-det}
  \vspace{-0.4cm}
\end{figure*}
}

\subsection{COCO-20$^i$}
\label{app:failure-cases}

{\begin{figure*}[h]
  \centering
  \includegraphics[width=0.7\linewidth]{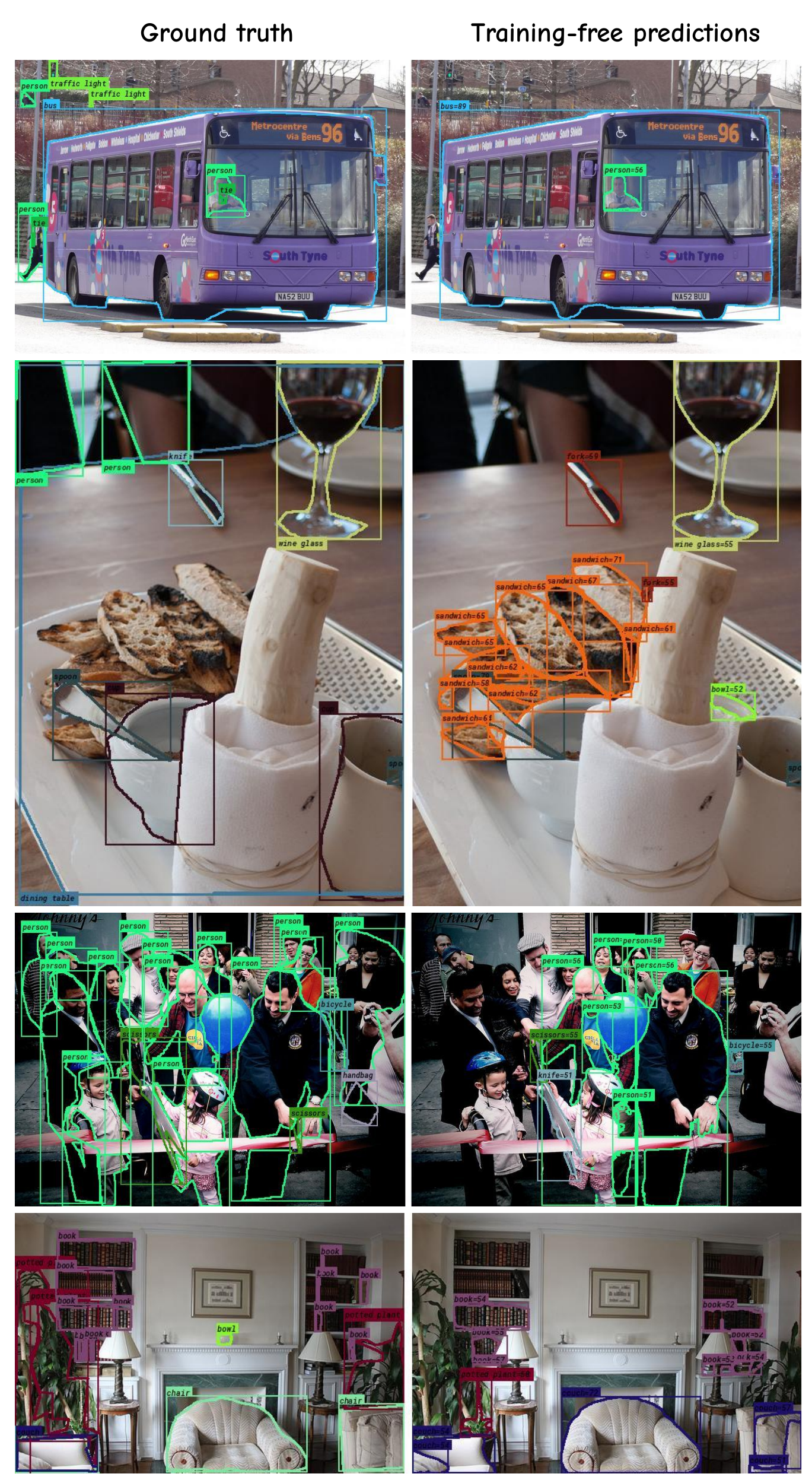}
  \caption{Visualisation of failure cases of our training-free method on the COCO val2017 set under the 10-shot setting (using 10 reference images per class). Our method sometimes confuses semantically similar classes, such as misclassifying bread as a hot dog or a large armchair as a couch. Additionally, we observe that fine or small objects are occasionally missed, and in highly crowded scenes, our model struggles to detect all instances accurately.}
  \label{fig:failure-cases}
\end{figure*}
}

Despite the strong performance of our training-free method across datasets, it also exhibits certain limitations, displayed in Figure \ref{fig:failure-cases}. A recurring failure mode is the confusion between semantically similar categories, such as bread being misidentified as a hot dog or large armchairs being mistaken for couches. This suggests that our approach could benefit from more fine-grained differentiation or improved selection of reference images. Additionally, detecting small or fine-grained objects remains challenging, as some instances are missed. Finally, in densely crowded scenes, where multiple overlapping objects appear, our model tends to under-detect instances, likely due to occlusions and the complexity of the visual context. These observations highlight areas for future improvement in robust object detection and segmentation under few-shot constraints.
}

\end{document}